\documentclass[conference]{IEEEtran}
\IEEEoverridecommandlockouts
\usepackage{cite}
\usepackage{amsmath,amssymb,amsfonts}
\usepackage{algorithmic}
\usepackage{graphicx}
\usepackage{textcomp}
\usepackage{xcolor}
\usepackage{xurl}
\usepackage{enumerate}
\usepackage{enumitem}
\usepackage{multirow}

\usepackage{pgfplotstable}
\usepackage{colortbl}
\def\BibTeX{{\rm B\kern-.05em{\sc i\kern-.025em b}\kern-.08em
    T\kern-.1667em\lower.7ex\hbox{E}\kern-.125emX}}

\pgfplotsset{compat=1.14}
\pgfplotstableset{
  color cells/.style={
    col sep=comma,
    string type,
    postproc cell content/.code={%
            \pgfkeysalso{@cell content=\rule{0cm}{2.4ex}\cellcolor{black!##1}\pgfmathtruncatemacro\number{##1}\ifnum\number>50\color{white}\fi##1}%
            },
    columns/x/.style={
        column name={},
        postproc cell content/.code={}
    }
  }
}
\begin{document}

\title{Learning Representations on Logs for AIOps
% \\
}

\author{\IEEEauthorblockN{Pranjal Gupta, Harshit Kumar, Debanjana Kar, Karan Bhukar, Pooja Aggarwal, Prateeti Mohapatra}
\IEEEauthorblockA{\textit{IBM Research, India} \\
% pranjal.gupta2@ibm.com, harshitk@in.ibm.com,  \{debanjana.kar1, karan.bhukar1\}@ibm.com, \\
% \{aggarwal.pooja, pramoh01\}@in.ibm.com
}}

\maketitle

\begin{abstract}
AI for IT Operations (AIOps) is a powerful platform that Site Reliability Engineers (SREs) use to automate and streamline operational workflows with minimal human intervention. Automated log analysis is a critical task in AIOps as it provides key insights for SREs to identify and address ongoing faults. Tasks such as log format detection, log classification, and log parsing are key components of automated log analysis. Most of these tasks require supervised learning; however, there are multiple challenges due to limited labeled log data and the diverse nature of log data. 
Large Language Models (LLMs) such as BERT and GPT3 are trained using self-supervision on a vast amount of unlabeled data. These models provide generalized representations that can be effectively used for various downstream tasks with limited labeled data. Motivated by the success of LLMs in specific domains like science and biology, this paper introduces a LLM for log data which is trained on public and proprietary log data. Results of our experiments demonstrate that the proposed LLM outperforms existing models on multiple downstream tasks. In summary, AIOps powered by LLMs offers an efficient and effective solution for automating log analysis tasks and enabling SREs to focus on higher-level tasks. Our proposed LLM, trained on public and proprietary log data, offers superior performance on multiple downstream tasks, making it a valuable addition to the AIOps platform.
\end{abstract}


\section{Introduction}
\label{sec:intro}
With the growing prevalence of scalable microservices-based applications, log analysis is an integral part of building robust systems\cite{loganalysis, loganalysis1}. As the scale of applications expands, the quantity of generated logs increases exponentially, posing a challenge for IT teams to analyze them manually\cite{mahindru2021log}. One critical aspect of building resilient systems is log analysis. By analyzing logs generated by various components of the system, IT teams can detect issues, identify their root cause(s), and take corrective action before they impact system availability. 
% Log analysis has emerged as an important task in both academia and industry \cite{loganalysis, loganalysis1} for the service reliability and diagnosis of microservices-based applications in cloud environment. It encompasses a range of tasks such as log format detection, error log detection, log parsing, and log anomaly detection.
In order to extract relevant information, log parsing is necessary, and log format detection plays a crucial role in this process. Accurately detecting the log format allows Site Reliability Engineers (SREs) to focus on the relevant logs and interpret log data effectively. After logs are parsed and collected, monitoring them is crucial for assessing the system's health. Logs can be categorized into different ``golden signals"~\cite{sre} to facilitate this monitoring process. The combination of log format detection and golden signal classification can reduce the mean time to detect an issue. Additionally, accurate fault category prediction is crucial in reducing the mean time to engage the right expert to handle the fault. A deeper understanding and representation of logs plays a crucial role in providing key insights to SREs in detecting faults, improving system availability, and minimizing downtime. 

One popular way of obtaining generalized representation for text is Large Language Models (LLMs). Recent years have seen the emergence of deep learning pre-trained LLMs such as BERT\cite{devlin-etal-2019-bert}, GPT-3\cite{gpt3}, PaLM\cite{peng-etal-2019-palm}, DALL-E\cite{dalle}, and Stable Diffusion\cite{rombach2021highresolution}. LLMs have been demonstrated to possess significant capabilities for representation learning and have the versatility to be utilized in diverse downstream applications, such as response generation, summarization, and text-to-image generation, etc. LLMs are built by infusing large amounts of data to train a deep neural network using one or more self-supervised auxiliary tasks. These models provide generalized representations that are not task-specific but can be fine-tuned for multiple downstream tasks, with limited labeled data in a few-shot setting. By leveraging the advanced representation learning capabilities of LLMs, log analysis can become even more efficient and effective, leading to significant improvements in application performance and reliability. Our focus is to utilize the log specific LLM on the three log analysis tasks: log format detection, golden signal classification, and fault category prediction.

\begin{table*}[tbh]
\centering
% \small
\caption{Sample log lines with annotations for the three downstream tasks}
\begin{tabular}{|p{5.25cm}|p{5.25cm}|lll|}
\hline
\multirow{2}{*}{\textbf{LogLine}} & \multirow{2}{*}{\textbf{Template}} & \multicolumn{3}{c|}{\textbf{Downstream Task Label}}                                            \\ \cline{3-5} 
                         &                           & \multicolumn{1}{l|}{\textbf{Log Format}} & \multicolumn{1}{l|}{\textbf{Golden Signal}} & \textbf{Fault Category} \\ \hline

% \multirow{LogLine} & \multirow{Template} & \multicolumn{3}{c|}{Downstream Task Label}                                            \\ \cline{3-5} 
%                          &                           & \multicolumn{1}{l|}{Log Format} & \multicolumn{1}{l|}{Golden Signal} & Fault Category \\ \hline

java.io.ioexception: Failed to create local dir in /opt/hdfs/nodemanager/usercache /curi/appcache/application  & java.io.IOException: Failed to create local dir in \textless*\textgreater                       & \multicolumn{1}{c|}{Spark}          & \multicolumn{1}{c|}{Error}             & \multicolumn{1}{c|}{I/O} \\ \hline

12-18 19:49:14.062 633 31868 e sdk: \textless2016-12-18 19:49:14\textgreater [ERR] SDK: UE-SeC 2016-12-18 19:49:14:62 Level[ERR] magic[3365677344 635]:TcpNonBlockReConnect failed because it can RW! err:110[Connection timed out]
                       & 12-18 \textless*\textgreater 633 31868 E SDK : \textless*\textgreater [ERR] SDK: UE-SeC 2016-12-18 \textless*\textgreater Level[ERR] magic[3365677344 635]:TcpNonBlockReConnect failed because it can RW! err:110[Connection timed out]                        & \multicolumn{1}{c|}{Android}          & \multicolumn{1}{c|}{Latency}             & \multicolumn{1}{c|}{Network} \\ \hline

% 2015-10-19 14:26:43,416 Fatal [IPC Server Handler 13 on 43581] org.apache.hadoop.mapred. TaskAttemptListenerImpl: Task: attempt\_1445182159119\_0003\_m\_00\_0 - failed due to FSError: java.io. IOException: There is not enough space on the disk
                    %   &  \textless*\textgreater FATAL [IPC Server handler \textless*\textgreater on \textless*\textgreater org.apache.hadoop.mapred. TaskAttemptListenerImpl: Task: \textless*\textgreater - failed due to FSError: java.io. IOException: There is not enough space on the disk  & \multicolumn{1}{c|}{Hadoop}          & \multicolumn{1}{c|}{Saturation}             & \multicolumn{1}{c|}{Memory} \\ \hline
                       
48587 node-235 action error 1076691608 1 clusterAddMember (cmd 2162) Error: Cannot run clusterAddMember\textbackslash because node-235 connection state is REFUSED, needs to be SRM
                       &  \textless*\textgreater \textless*\textgreater action error \textless*\textgreater 1 clusterAddMember (cmd \textless*\textgreater Error: Cannot run clusterAddMember\textbackslash because \textless*\textgreater connection state is REFUSED, needs to be SRM & \multicolumn{1}{c|}{HPC}          & \multicolumn{1}{c|}{Availability}             & \multicolumn{1}{c|}{Network} \\ \hline
\end{tabular}
\label{tab:sample-annotations}
\end{table*}
Directly applying existing LLMs, which are pre-trained on natural language text, to semi-structured log data can be challenging due to the dissimilarities between the two data types. For instance, the lexicon and vocabulary of log data differs from natural language text; this is because log data contains domain-specific words and phrases that are not commonly used outside of their respective fields. As most log messages have a limited grammatical structure, the representations or embeddings provided by pre-trained LLMs do not provide the best results on downstream tasks. Also, capturing token positional information in logs is crucial as they help in log parsing tasks. Displayed in Table~\ref{tab:sample-annotations} are examples of log lines, their corresponding templatized forms\cite{10.1109/ICSE-SEIP.2019.00021}, and the labels assigned to them based on their downstream tasks. 
% (Note: A \textit{template} refers to a representation of a print statement in source code that appears with different parameter values in multiple runs, resulting in varying raw log outputs. ~\cite{10.1109/ICSE-SEIP.2019.00021}). 
These examples demonstrate the crucial role of comprehending the inherent structure, vocabulary, and attributes of log messages in facilitating efficient log analysis.
% Table \ref{tab:sample-annotations} shows sample log lines, their templatized forms (A \textit{template} is a representation of a print statement in source code that appears with different parameter values in multiple runs, resulting in varying raw log outputs. ~\cite{10.1109/ICSE-SEIP.2019.00021}), and their associated labels depending on the downstream tasks. 
% As can be seen from these examples, understanding inherent structure, vocabulary and characteristics of log messages is key for efficient log analysis. %Since log data is semi-structured, it can be represented as a set of templates. Even after templatization, it contains long sentences that include error messages, concatenated tokens with camel-case formatting, and abbreviations. 
Similar observations have also been reported by several other prior works \cite{10.1093/bioinformatics/btz682,chalkidis2020legal,alsentzer2019publicly,zhang2020rapid} that pre-trained LLMs do not perform as expected on domain-specific tasks.

Inspired by the LLMs for different domains, such as TweetBERT \cite{nguyen-etal-2020-bertweet}, SciBERT \cite{beltagy-etal-2019-scibert}, and BioBERT \cite{10.1093/bioinformatics/btz682}, we introduce \textit{BERTOps}, an LLM for AI in operations (AIOps) domain, pre-trained over large-scale public and proprietary log data. The emergence of domain-specific LLMs is due to the fact that existing LLMs are trained on general corpora such as news articles and Wikipedia, and hence do not perform as expected for specific domains. To the best of our knowledge, \textit{BERTOps} is the first LLM that can effectively
generalize to multiple downstream tasks of log analysis. We finetune the pre-trained \textit{BERTOps} model under a few-shot setting for three downstream tasks - Log Format Detection (LFD), Golden Signal Classification (GSC), and Fault Category Prediction (FCP). For the aforementioned tasks, we show that BERTOps outperforms classical machine learning models and LLMs. The main contributions of this paper are as follows:
\begin{enumerate}[leftmargin=*,align=left]
\item We propose an encoder-based LLM (\textit{BERTOps}) for the AIOps domain, and show its application on the three downstream tasks for log analysis - Log Format Detection, Golden Signal Classification, and Fault Category Prediction.

\item We provide labeled data for the aforementioned three downstream tasks, and release it publicly along with the code base.  We believe that these tasks along with the datasets and the model will work as a benchmark for further research. To the best of our knowledge, there are currently no existing labeled datasets available for the Golden Signal Classification task and Fault Category Prediction task.

\item Our experiments suggest that the encoder-based \textit{BERTOps} exhibit few-shot generalization with purely unsupervised training. To demonstrate the effectiveness of LLM for AIOps, we compare the \textit{BERTOps} with classical machine learning models and pre-trained encoder-based models such as BERT and RoBERTa. An important observation is a significant gain in performance of \textit{BERTOps} visa-a-vis classical ML models for log analysis tasks.

\end{enumerate}
% For the aforementioned tasks, we show that BERTOps performs well in comparison to pre-trained BERT models and classical machine learning models.  

%For instance, one such influential LLM in the language domain is Bidirectional Encoder Representations from Transformers (BERT) \cite{devlin-etal-2019-bert}. It was pre-trained on BookWiki Corpus - consisting of English Wikipedia and BooksCorups - with two auxiliary tasks Masked Language Model (MLM) and Next Sentence Prediction (NSP), to capture the language capability of natural English language. And, it has been shown remarkable performance on several downstream tasks ranging from question answering, natural language inference, classification, named-entity recognition, etc. 

\begin{figure*}[tbh]
    \centering
    \includegraphics[scale=0.35]{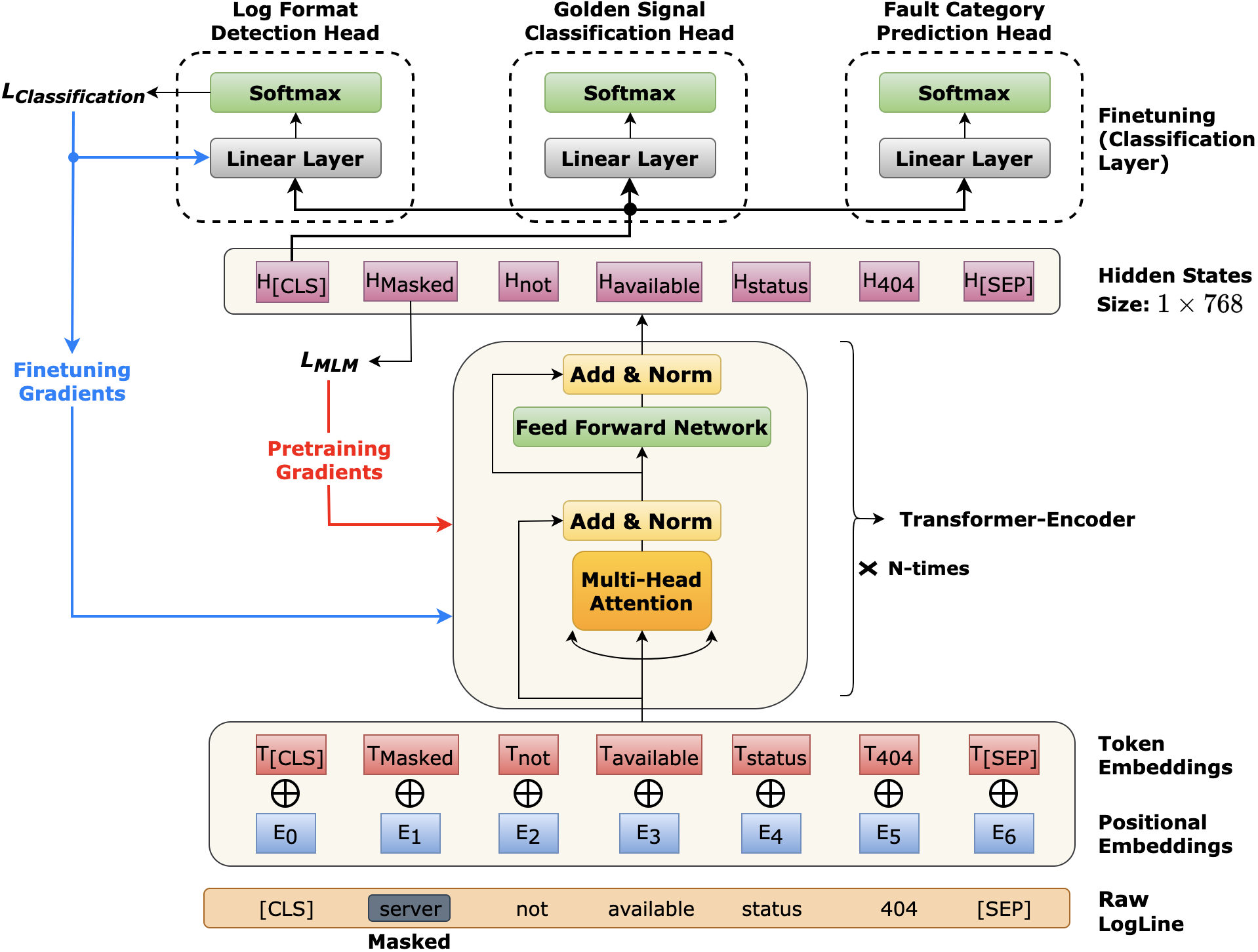}
    % \caption {Overall pre-training and fine-tuning process of BERTOps. During Pretraining transformer encoder is trained using the masked LM ($L_{MLM}$) of BERT. To fine-tune pretrained BERTOps on a given downstream task, a downstream task's $L_{classification}$ classification loss is used. The unique $[CLS]$ token, which captures the entire semantics of the raw log line, is used as the classification layer input.} 
    \caption{BERTOPs end-to-end architecture pre-training and finetuning on three downstream tasks. The model has been designed to be plug-and-play such that new downstream tasks can be added when required. It computes two losses: pretraining loss $L_{MLM}$ and task specific finetuning loss $L_{classification}$.}
    \label{fig:model}
\end{figure*}
\section{Method}
\label{sec:method}
\begin{figure}[tbh]
    \centering
    \includegraphics[width=0.48\textwidth, trim=0cm 0.2cm 0cm 1.3cm, clip]{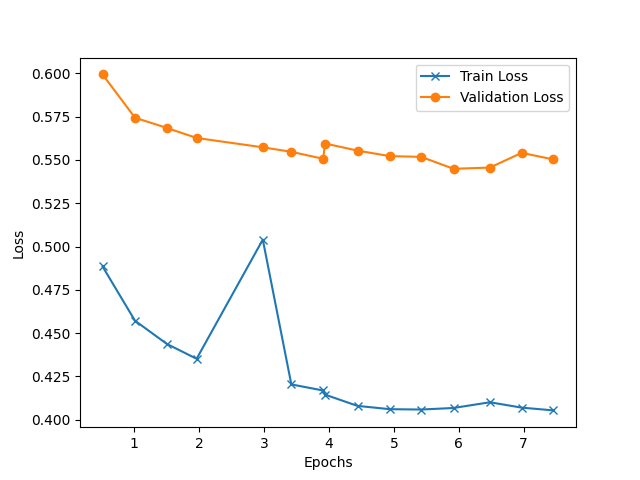}
    \caption{Train vs Validation Loss during pretraining of BERTOps}
    \label{fig:train-vs-val-loss}
\end{figure}

%\subsection{Background}
In this section, we introduce \textit{BERTOps}, a large-scale Transformer model for log analysis and understanding. Figure~\ref{fig:model} shows an end-to-end architecture of the \textit{BERTOps}, including pretraining the model and finetuning it on the downstream tasks. The downstream tasks for log analysis are mostly classification-based, therefore we build an encoder-based LLMs for log data. The architecture design of \textit{BERTOps} is motivated from \textit{BERT-BASE}\cite{devlin-etal-2019-bert} with $12$ layers, $12$ self-attention heads, and the size of representation layer is $768$. Since there is a vocabulary overlap between natural text and log data, especially for non domain-specific words and phrases, the pretrained weights from BERT-BASE are used as the initialization weights for \textit{BERTOps} and further pretrain it on logs data. The intuition is to bootstrap \textit{BERTOps} with BERT's knowledge of natural language text, so that it can focus more on understanding log data specific vocabulary and
% so that BERTOps doesn't need to reinvent the wheel from scratch,
representation during pretraining. The transformer encoder of BERTOps is further pretrained on logs data using the masked language modeling (MLM) task\cite{devlin-etal-2019-bert}. During pretraining phase, some percentage of tokens in a log sequence are randomly masked as shown in Figure~\ref{fig:model}. The objective of the MLM task is to predict the token corresponding to the masked token based on the neighbouring context tokens. For example, Figure~\ref{fig:model} demonstrates an instance of token masking in which the token "server" is masked. The objective of the MLM task is to utilize the embedding $H_{masked}$ to predict the original token, "server", by employing cross-entropy loss ($L_{MLM}$). We continue pretraining the BERTOps until training loss($L_{MLM}$) saturates, and use the model with the least validation loss as the final model. For example,  Figure~\ref{fig:train-vs-val-loss} shows that at epoch $5$, the validation loss is minimum. Also, note that the validation loss saturates between epochs $5$ and $6$. Although we continued training after epoch $6$, hoping that loss would further reduce, however, that did not happen. Instead, it went up a little and in the following epoch, it came back to the same level which is a sign that the loss is converging. Therefore, we used the model checkpoint at epoch $6$ as the final model for the downstream tasks. 
Besides using validation loss for model selection, we also used perplexity scores to monitor the progress during pretraining. Perplexity~\cite{perplexity} measures how well a LLM predicts a sequence of words. The lower the perplexity score, the better the language model is at predicting the next word. It is calculated as follows:
\begin{equation*}
\small
PP(\mathbf{w}) = \sqrt[n]{\frac{1}{P(w_1, w_2, ..., w_N)}}
\end{equation*}
where $\mathbf{w} = {w_1, w_2, ..., w_N}$ is a sequence of words, and $P(w_1, w_2, ..., w_N)$ is the joint probability of the sequence of words to occur together.
As the training continued, the perplexity score came down from $87.9937$ at epoch $0$ to $1.7363$ at epoch $5.4$, the same checkpoint which had the least validation loss. The high perplexity score in epoch $0$ validates the hypothesis mentioned in the Introduction (Section~\ref{sec:intro}) that log lines do not have proper grammatical structure as opposed to natural language text. On further training, the perplexity score dropped, this suggests that the \textit{BERTOps} adjusted its weights to build representations in accordance with the syntactical and semantical structure of words appearing in log data. 
% While training the model, the stopping criteria is until the training loss saturation for few epochs and choose the checkpoint with the least validation loss as the final model for the downstream tasks.
\begin{table}[tbh]
\centering
% \small
\caption{"Public" Pretraining dataset statistics}
% \resizebox{\columnwidth}{!}{%
\begin{tabular}{|l|p{1cm}|p{1.3cm}|p{1.3cm}|p{1.4cm}|}
\hline
\textbf{LogSource} & \textbf{Train} & \textbf{Validation} & \textbf{\#Instances} & \textbf{\#Templates} \\ \hline
Android              & 1244K             & 311K                 & 1555K                & 5,852                 \\ \hline
BGL                  & 3798K             & 949K                 & 4747K                & 584                   \\ \hline
HDFS                 & 8940K             & 2235K               & 11175K               & 95                    \\ \hline
HPC                  & 346K               & 86K                  & 433K                  & 208                   \\ \hline
Hadoop               & 315K               & 78K                  & 394K                  & 475                   \\ \hline
Linux                & 204K                & 5K                   & 25K                   & 196                   \\ \hline
Mac                  & 93K                & 23K                  & 117K                  & 1188                 \\ \hline
Openstack            & 151K               & 37K                  & 189K                  & 158                   \\ \hline
Proxifier            & 17K                & 4K                   & 21K                   & 82                    \\ \hline
Spark                & 26589K            & 6647K               & 33236K               & 170                   \\ \hline
Thunderbird          & 80K                & 20K                  & 100K                  & 1465                 \\ \hline
Zookeeper            & 59K                & 14K                  & 74K                   & 61                    \\ \hline
\textbf{Total}       & \textbf{41.6M}   & \textbf{10.4M}     & \textbf{52.07M}      & \textbf{10.5K}       \\ \hline
\end{tabular}%
% }
\label{tab:public-data-stats}
\end{table}

\begin{table}[tbh]
\centering
% \small
% \resizebox{\columnwidth}{!}{%
\caption{"Proprietary" Pretraining dataset statistics}
\begin{tabular}{|l|p{1cm}|p{1.3cm}|p{1.3cm}|p{1.4cm}|}
\hline
\textbf{LogSource} & \textbf{Train} & \textbf{Validation} & \textbf{\#Instances} & \textbf{\#Templates} \\ \hline
Apache              & 1719K             & 430K                 & 2149K                & 372                   \\ \hline
HAProxy             & 4K                 & 1K                   & 5K                    & 6                     \\ \hline
MongoDB             & 120K               & 30K                  & 150K                  & 14                    \\ \hline
Roboshop            & 17K                & 4K                   & 21K                   & 206                   \\ \hline
Sockshop            & 87K                & 22K                  & 108K                  & 179                   \\ \hline
\textbf{Total}      & \textbf{1.9M}    & \textbf{487K}        & \textbf{2435K}       & \textbf{777}          \\ \hline
\end{tabular}%
% }
\label{tab:proprietary-dataset-stats}
\end{table}
As mentioned earlier, pretraining LLMs requires a huge amount of data. For BERTOps, the pretraining data consists of logs from $17$ different data sources, out of which $12$ are collected from an open-source repository\cite{he_loghub_2020}, and the remaining $5$ are proprietary data sources. Tables~\ref{tab:public-data-stats} and~\ref{tab:proprietary-dataset-stats} show the number of instances per log format for pretraining BERTOps. Proprietary data sources consist of log formats from custom applications whose structure are different from existing public log sources. 
The proprietary data sources only constitute $~0.937 \%$ of the entire pretraining corpus. Therefore, we can infer that using only public datasets for pretraining shouldn't have a significant deviation from the results presented in this paper. In order to generate the training and validation sets for our experiments, an $80:20$ split was performed on each log source. Subsequently, the resulting data was combined to form the training and validation sets, which comprised of $43.5M$ and $10.9M$ log lines, respectively.
% \usepackage{graphicx}
% \usepackage{graphicx}

Data preprocessing of pretraining datasets consists of splitting tokens that are in camelcase format or tokens joined with periods or dash using regular expressions followed by converting tokens to lower case. The pretraining for BERTOps involved 8 epochs that spanned across a 20-day period. Huggingface's implementation\footnote{\url{https://github.com/huggingface/transformers/blob/main/examples/pytorch/language-modeling/run\_mlm.py}} was used for pretraining BERTOps on $4$ $A100$ GPUS with a batch size of $256$. 
% We use Huggingface's implementation\footnote{\url{https://github.com/huggingface/transformers/blob/main/examples/pytorch/language-modeling/run\_mlm.py}} for pretraining BERTOps. 
% The data preprocessing steps involve using regular expressions to split phrases in camelcase format, joined with periods or dash. Furthermore,  convert all the resulting tokens to lowercase. 
To obtain statistics for \textit{Out-of-Vocabulary (OOV)} words in the pretraining dataset, we calculated the frequency of $<$unk$>$ tokens (which signify OOV). Our analysis revealed that after utilizing BERT's tokenizer, $0.0062\%$ and $0.0061\%$ of tokens in the training and validation data, respectively, correspond to OOV. Once pretraining of BERTOps model is complete, it is finetuned using a cross-entropy-loss $L_{classification}$ for each of the task separately. More details for finetuning BERTOps are discussed in subsection \ref{sec:finetuning}
\section{Experimental Setup}
The usefulness of a pretrained LLM hinges upon how well it performs across multiple downstream tasks. That is, how effectively can it apply or transfer the knowledge it gained through pre-training on large corpuses for the downstream tasks. This section introduces three downstream tasks for log analysis in AIOps domain, process of preparing labeled datasets for these tasks, and finetuning the pretrained LLM on these tasks. The three tasks are: Log Format Detection, Golden Signal Classification, and Fault Category Prediction. 

\subsection{Downstream Tasks}
\label{sec:downstream-tasks}
To evaluate the effectiveness of the \textit{BERTOps} model and the other baselines, three downstream log analysis tasks are defined.
\begin{enumerate}[leftmargin=*,align=left]
    \item {\textbf{Log Format Detection (LFD)}}  Identifying the format of logs is an important first step of any log analysis pipeline \cite{4601543,5936060,8973030}, which can help leverage the knowledge of its unique structures to parse the logs and aid in tasks like structure extraction \cite{he_drain_2017}, key entity/metric extraction, anomaly detection \cite{yang2021semi}, etc. With multiple log variations within each log format (refer column Templates in Table \ref{tab:lfd-data}), learning to distinguish them from a few training samples makes it a challenging task. In this task, given logs from varied sources, we train a multi-class classification model that learns to distinguish logs from $16$ different log formats. Tables  \ref{tab:public-data-stats} and \ref{tab:proprietary-dataset-stats} provides the number of templates (indicating variations) per format.
\begin{table}[tbh]
\centering
% \small
\caption{Sample Log Variations in SSH and HDFS Log formats}
\begin{tabular}{|c|p{7cm}|}
\hline
\textbf{Format}             & \textbf{Template} \\ \hline
\multirow{3}{*}{HDFS} & \textless{}*\textgreater INFO dfs.DataNode\$PacketResponder:  PacketResponder 2 for block\textless{}*\textgreater terminating        \\ \cline{2-2} 
                   & \textless{}*\textgreater INFO dfs.FSNamesystem: BLOCK* NameSystem.allocateBlock: \textless*\textgreater \textless*\textgreater       \\ \cline{2-2} 
                   & \textless*\textgreater INFO dfs.DataNode\$BlockReceiver: Exception in receiveBlock for block \textless{}*\textgreater java.io.IOException: Connection reset by peer       \\ \hline
\multirow{3}{*}{SSH} & \textless*\textgreater LabSZ \textless*\textgreater Failed none for invalid user \textless*\textgreater from \textless*\textgreater port \textless*\textgreater ssh2        \\ \cline{2-2} 
                   & \textless*\textgreater LabSZ \textless*\textgreater Invalid user \textless*\textgreater from \textless*\textgreater       \\ \cline{2-2} 
                   & \textless{}*\textgreater LabSZ \textless{}*\textgreater Received disconnect from \textless{}*\textgreater 11: Bye Bye {[}preauth{]}         \\ \hline
\end{tabular}
\label{tab:lfd-data}
\end{table}
    \item {\textbf{Golden Signal Classification (GSC)}} Google SRE handbook for SREs\cite{sre} outlines basic principles and practices for monitoring applications hosted on cloud. 
% Golden Signal are very useful for monitoring events and alerts emanating from logs and metrics generated by an application hosted on cloud. 
Golden Signals are a set of key performance indicators (KPIs) used for monitoring log and metric events to analyze health of a system. One of the key benefits of monitoring the golden signals is that it offers a convenient and efficient method for detecting and troubleshooting errors. By monitoring these KPIs, it is possible to detect anomalies and trends that can indicate problems before they become critical. This enables teams to proactively respond to issues and ensure that their systems are running smoothly and efficiently. The four golden signals defined in the SRE handbook are latency, traffic, errors, and saturation. Latency measures the time it takes for a request to be processed, errors count the number of failed requests, saturation measures the degree to which system resources are being used, and traffic measure of how much demand is being placed on your system. 
\begin{table}[tbh]
\centering
% \small
\caption{Log examples with golden signal labels}
\begin{tabular}{|l|c|}
\hline
\multicolumn{1}{|c|}{\textbf{LogLine}}                                                                                                              & \textbf{Golden Signal} \\ \hline
\begin{tabular}[c]{@{}l@{}}java.net.NoRouteToHostException: \\ No route to host: no further information\end{tabular}                              & Availability           \\ \hline
\begin{tabular}[c]{@{}l@{}}DateTime Got an XPC error: Connection \\ invalid\end{tabular}                                                          & Error                  \\ \hline
DateTime INFO Notification time out 60000                                                                                                         & Latency                \\ \hline
\begin{tabular}[c]{@{}l@{}}Caused by: org.apache.hadoop.fs.FSError:\\  java.io.IOException: There is not enough\\  space on the disk\end{tabular} & Saturation             \\ \hline
\begin{tabular}[c]{@{}l@{}}SohuNews.exe - \_URL\_ close, 673\\  bytes sent, 384 bytes received, lifetime 00:18\end{tabular}                       & Traffic                \\ \hline
\begin{tabular}[c]{@{}l@{}}12-18 15:33:40.463 3180 3578 i ash :\\ application com.tencent.mm is im app and\\ connected ok\end{tabular}                       & Information                \\ \hline
\end{tabular}
\label{tab:gs_examples}
\end{table}
% The four golden signals defined in the SRE handbook are latency, traffic, errors, and saturation. Latency measures the time it takes for a request to be processed, traffic measures the volume of requests, errors count the number of failed requests, and saturation measures the degree to which system resources are being used.
We introduce "availability", a new golden signal to the set of golden signals. The intuition to introduce a new class is to capture a special type of error that needs immediate SRE attention. When a logline indicates an availability golden signal, it means that the underlying application is not responding or device is not reachable and may lead to a critical fault. Also, there are certain log messages that are non-erroneous, such as heart beat messages, trace messages, etc - such log messages do not belong to the existing golden signals, we have defined an additional class for them called as "Information".  
Table \ref{tab:gs_examples} enumerates examples of each golden signal classification label.

    \item {\textbf{Fault Category Prediction (FCP)}} Fault categories associated with logs serve as signals to detect anomalous system behaviours and provide clues for failure diagnosis. That is, a Fault category helps to understand why a service fails and may help in isolating the root causes of the failure. In addition, fault categories can also be used to route the ongoing fault/issue to appropriate teams for debugging and remediation. Through this task, we aim to build a fault categorization model that classifies a log line or raw log message into one of the $7$ fault categories: Memory, Network, Authentication, I/O, Device, Application, and Other\cite{fault_category}.
    % To build a fault categorization model, we train our model to classify logs into possible fault categories.  %resulting in a multi label distribution. 
    % We consider $7$ fault categories (memory, network, authentication, i/o, application, device, other) as proposed in \cite{fault_category}. 
    Table \ref{tab:fc} provides an example for each fault category.
\end{enumerate}
\begin{table}[tbh]
\centering
% \small
\caption{Log examples with fault category labels}
\begin{tabular}{|p{6cm}|p{1.8cm}|}
\hline
\textbf{LogLine} & \textbf{Fault Category} \\ \hline
WARN dfs.FSDataset: Unexpected error trying to delete block blk 6566051927569845875. BlockInfonot found in volumeMap. & Memory \\ \hline

Connection not available while invoking method createOrWaitForConnection for resource jdbc/Actil & Network \\ \hline

SSL HANDSHAKE FAILURE: A signer with SubjectDN was sent from a target host. The certificate expired at Mon Feb 15 & Authentication  \\ \hline

WARN [ResponseProcessor for block: Slow ReadProcessor read fields took 48944ms (threshold=30000ms); SUCCESS status: &  I/O \\ \hline

Server launched but failed initialization. Server logs,startServer.log, and other log files under. & Device \\ \hline
An exception was thrown by one of the service methodsof the servlet in application & Application \\ \hline

081110 065758 7079 INFO dfs.DataNode\$DataXceiver: 10.251.106.10:50010 Served block blk\_-6040371292695818218 to 10.251.74.134 & Other \\ \hline
\end{tabular}
\label{tab:fc}
\end{table}

\subsection{Labeled Datasets for Downstream Tasks}
\label{sec:dataset-prep}
Training a model for a task from scratch requires huge amount of labeled data instances which are hard and expensive to generate. Since we aim to evaluate the effectiveness of the fine-tuned LLM for the downstream tasks with only a few labeled training examples, it is crucial to have an extensive labeled test set to validate its performance.
% Since, we want to evaluate how effective is an LLM for the downstream tasks when fine-tuned with a few labelled training examples, we also need an extensive labeled test set to validate its performance. 
This work aims to provide a labeled training dataset for fine-tuning models under a few-shot setting. Additionally, an annotated test dataset will be made available, which can serve as a suitable benchmark for these tasks, thereby enabling researchers to make comparisons and evaluate the performance of their models. Log data for the downstream tasks is sourced from LogHub \cite{he_loghub_2020} and proprietary data sources. While we had access to gold-standard annotations for the LFD task, we manually curated annotated data samples for the other two tasks. 
% We plan to release these datasets for general availability post publication.

For the LFD task, a dataset was prepared from the $16$ formats consisting of Android, Apache, BGL, HDFS, HPC, Hadoop, HealthApp, Mac, Openstack, Proxifier, SSH, Syslog-Sendmail, Spark, Thunderbird, Websphere and Zookeeper. Out of the $16$ log formats, the 
$4$ log formats (HealthApp, SSH, Sendmail, WebSphere) were treated as held-out, i.e., these $4$ log formats were not used for pretraining \textit{BERTOps} (refer Table ~\ref{tab:public-data-stats} and ~\ref{tab:proprietary-dataset-stats}), but they will be required during finetuning the pretrained LLM models for the downstream tasks. The goal of holding out these $4$ datasets is to test the generalizability of LLMs, including \textit{BERTOps}, to handle unseen datasets for the downstream tasks.

Preparing labeled datasets for the GS Classification and the FC Predication tasks is a challenging process for two reasons: one, the amount of data is huge, and second, it is practically infeasible for human annotators to label each logline. To mitigate this situation, we templatized logs for each format using the Drain template miner~\cite{he_drain_2017}. Templatization allows to cluster logs into homogeneous groups that have the same structure. We distributed few instance of each log template among the annotators. For example, Table~\ref{tab:proprietary-dataset-stats} shows that the number of log lines for MongoDB log format are approximately 150K which can be grouped into 14 templates. Instead of labeling each of the 150K log lines, the annotators were asked to label only 14 templates. The template labels were reverse mapped to actual log lines associated with each template. This reduced the human effort of manually labeling each log line by manifold while getting a good coverage.

% In-order to prepare labeled datasets for the Golden Signal Classification (GS) and the Fault Category Prediction (FC) tasks, we templatized the logs from each format using the Drain template miner~\cite{8029742} to detect variations among them. Then we grouped logs by each log template. Furthermore, we manually labeled each log template manually, and propagated the assigned label for each log template format to its associated group of instances.  
% For each of the tasks, we prepared challenging datasets by carefully choosing log formats that contain a lot of variations.
% \todo{add justification for excluding traffic class}
For GS classification task, $7$ subject matter experts were provided with $272$ templates for labeling. The overlap of log templates among the annotators was $66\%$. Similarly, for FC prediction task, $394$ templates were provided for labeling with an overlap of $43\%$. After labeling, it was observed that the Traffic class had a skewed distribution with very few examples. This skewed distribution ultimately diminished the overall quality of the dataset. As a result, the final labeled dataset does not include the Traffic class.
%After labeling, we observed that the Traffic class has a skewed distribution with very few examples, ultimately diminishing the overall quality of dataset, therefore the final labeled dataset does not include Traffic class. 
Once the labeling of templates for GS and FC tasks was completed, those templates where annotators had disagreements were revisited to resolve the differences. During the disagreement resolution process, conflicts were resolved using majority vote. Kappa coefficient \cite{kappa} is used as a metric for computing inter-annotator agreement. For the GSC and FCP, the inter-annotator agreement is $60.62$ and $65.80$, respectively. Inter-annotator agreement in the range of 60-65 indicates the complexity associated with these tasks.

To collate the k-shot training dataset for each task, where $k \in \{10, 20, 30\}$, we randomly selected $k$ templates with discernible variations. We curate 10, 20 and 30 shot training datasets for each task by finding 10, 20, 30 samples of each label for each task. For example, 10-shot training set for log format detection will have $16 \times 10$ log samples. We used the rest of the templates along with their group of instances as the test set. For each of the three downstream tasks, the data statistics of the labeled dataset are presented in  Table \ref{tab:dataset}.

% After the templates were labelled, the annotator disagreements were resolved by taking a majority vote among the annotators. We templatize the logs from each format to detect variations among them. For each of the tasks, we have prepared a challenging dataset by carefully choosing log formats which contain a lot of variations.  To collate the k-shot training dataset for each task, where $k \in {10, 20, 30}$, we manually choose $k$ templates with discernible variations and take one log example corresponding to each template. We use the rest of the templates as our test set. We present our data distribution in Table \ref{tab:dataset}.

% \usepackage{multirow}
\begin{table}[tbh]
% \small
\centering
\caption{Downstream tasks data distribution}
\begin{tabular}{|l|c|c|}
\hline
\multicolumn{1}{|c|}{\textbf{Task}}   & \textbf{\# Classes} & \textbf{\# Test} \\ \hline
Log Format Detection         & 16         & 11918   \\
Golden Signal Classification & 5          & 423     \\
Fault Category Prediction    & 7          & 868     \\ \hline
\end{tabular}
% We curate 10, 20 and 30 shot training datasets for each task by finding 10, 20, 30 samples of each label for each task (for eg. , 10-shot training set for log format detection will have 16X10 log samples).
\label{tab:dataset}
\end{table}

\subsection{Finetuning}
\label{sec:finetuning}
A widely used industry practice is to build an LLM which upon release is adopted and deployed for the downstream tasks by application-focused developers and businesses. The pre-trained \textit{BERTOps} model prepared in Section~\ref{sec:method} is further finetuned for each task separately. Finetuning a model for a particular task requires labelled data. In the real-world scenario, clients find it difficult to share large data payloads with private data to train models for various tasks. The other challenge is the availability of labeled data. Moreover, data annotation for each task is an expensive and laborious exercise. Few-shot learning mitigates that bottleneck by teaching the models to classify with very few data samples. 

For the downstream tasks described in Section~\ref{sec:downstream-tasks}, we finetune BERTOps in a few-shot setting to evaluate the effectiveness of our learned transformer representations. As shown in Figure~\ref{fig:model} we add a classification layer on top of BERTOps consisting of a linear layer followed by softmax activation. We use simpletransformers\footnote{\url{https://simpletransformers.ai/}} library for performing experiments on downstream tasks. All the downstream tasks use Cross Entropy loss as the objective function. The finetuning training process updates all model parameters including the pretrained weights. These experiments are performed on a single $A100$ GPU server for $20$ epochs with AdamW optimizer \cite{AdamW} and learning rate of $4e\text{-}5$. 
% The code will be made available post publication.

% For each of the above described downstream tasks, we perform few shot experiments to evaluate the effectiveness of our learned transformer representations. We have used a Linear layer classifier followed by a sigmoid activation to finetune the representations on the above mentioned downstream tasks. We performed these experiments on a single A100 GPU server for $20$ epochs. 

\section{Experiments}
In this section, we present results of our experimental study, followed by its analysis. The experimental studies are designed to answer the following questions: (1) Is an AIOps domain specific LLM required? (2) How well does the proposed \textit{BERTOps} LLM performs in comparison to the other baseline models for the three downstream tasks, i.e. Log Format Detection, Golden Signal Classification, and Fault Category Prediction tasks? (3) How effective is the few-shot learning approach?
%, and is it possible to expand its usage to other tasks in AIOPs? \todo{PA: fewshot -> to include in the introduction about the model can be repurposed/finetuned for a new downstream task}

% \usepackage{multirow}
%=======================================================================
\begin{table*}[tbh]
\centering
% \small
\caption{Comparing baselines with BERTOps on Log Format Detection}
\begin{tabular}{|clccccccccc|}
\hline
\multicolumn{11}{|c|}{Log Format Detection} \\ \hline
\multicolumn{1}{|c|}{\multirow{2}{*}{Model Type}} & \multicolumn{1}{c|}{\multirow{2}{*}{Model}} & \multicolumn{3}{c|}{10-shot} & \multicolumn{3}{c|}{20-shot}  & \multicolumn{3}{c|}{30-shot} \\ \cline{3-11} 
\multicolumn{1}{|c|}{}                                      & \multicolumn{1}{c|}{}                       & \multicolumn{1}{c|}{P}              & \multicolumn{1}{c|}{R}              & \multicolumn{1}{c|}{F1}             & \multicolumn{1}{c|}{P}              & \multicolumn{1}{c|}{R}              & \multicolumn{1}{c|}{F1}             & \multicolumn{1}{c|}{P}              & \multicolumn{1}{c|}{R}              & F1             \\ \hline
\multicolumn{1}{|c|}{\multirow{2}{*}{Classical ML}}         & \multicolumn{1}{l|}{Decision Tree}          & \multicolumn{1}{c|}{93.49}          & \multicolumn{1}{c|}{71.86}          & \multicolumn{1}{c|}{75.10}          & \multicolumn{1}{c|}{93.05}          & \multicolumn{1}{c|}{77.57}          & \multicolumn{1}{c|}{81.35}          & \multicolumn{1}{c|}{92.08}          & \multicolumn{1}{c|}{83.34}          & 85.08          \\ \cline{2-11} 
\multicolumn{1}{|c|}{}                                      & \multicolumn{1}{l|}{SGD}                    & \multicolumn{1}{c|}{89.10}          & \multicolumn{1}{c|}{87.15}          & \multicolumn{1}{c|}{87.17}          & \multicolumn{1}{c|}{90.21}          & \multicolumn{1}{c|}{87.98}          & \multicolumn{1}{c|}{88.04}          & \multicolumn{1}{c|}{91.87}          & \multicolumn{1}{c|}{90.35}          & 90.34          \\ \hline
\multicolumn{1}{|c|}{\multirow{5}{*}{Pretrained LLMs}}      & \multicolumn{1}{l|}{ALBERT-base-v2}         & \multicolumn{1}{c|}{95.54}          & \multicolumn{1}{c|}{94.26}          & \multicolumn{1}{c|}{94.58}          & \multicolumn{1}{c|}{96.79}          & \multicolumn{1}{c|}{96.27}          & \multicolumn{1}{c|}{96.40}          & \multicolumn{1}{c|}{98.20}          & \multicolumn{1}{c|}{98.19}          & 98.19          \\ \cline{2-11} 
\multicolumn{1}{|c|}{}                                      & \multicolumn{1}{l|}{ELECTRA}                & \multicolumn{1}{c|}{89.53}          & \multicolumn{1}{c|}{85.87}          & \multicolumn{1}{c|}{86.24}          & \multicolumn{1}{c|}{95.67}          & \multicolumn{1}{c|}{94.74}          & \multicolumn{1}{c|}{94.94}          & \multicolumn{1}{c|}{98.03}          & \multicolumn{1}{c|}{97.97}          & 97.98          \\ \cline{2-11} 
\multicolumn{1}{|c|}{}                                      & \multicolumn{1}{l|}{XLNet}                  & \multicolumn{1}{c|}{95.34}          & \multicolumn{1}{c|}{94.76}          & \multicolumn{1}{c|}{94.83}          & \multicolumn{1}{c|}{96.41}          & \multicolumn{1}{c|}{96.24}          & \multicolumn{1}{c|}{96.27}          & \multicolumn{1}{c|}{97.96}          & \multicolumn{1}{c|}{97.77}          & 97.81          \\ \cline{2-11} 
\multicolumn{1}{|c|}{}                                      & \multicolumn{1}{l|}{RoBERTa}                & \multicolumn{1}{c|}{95.17}          & \multicolumn{1}{c|}{94.24}          & \multicolumn{1}{c|}{94.42}          & \multicolumn{1}{c|}{96.72}          & \multicolumn{1}{c|}{96.49}          & \multicolumn{1}{c|}{96.55}          & \multicolumn{1}{c|}{98.31}          & \multicolumn{1}{c|}{98.30}          & 98.30          \\ \cline{2-11} 
\multicolumn{1}{|c|}{}                                      & \multicolumn{1}{l|}{BERT-base}              & \multicolumn{1}{c|}{95.61}          & \multicolumn{1}{c|}{95.17}          & \multicolumn{1}{c|}{95.23}          & \multicolumn{1}{c|}{96.96}          & \multicolumn{1}{c|}{96.81}          & \multicolumn{1}{c|}{96.84}          & \multicolumn{1}{c|}{97.63}          & \multicolumn{1}{c|}{97.42}          & 97.47          \\ \hline
\multicolumn{1}{|c|}{\textbf{Proposed Domain Specific LLM}} & \multicolumn{1}{l|}{\textbf{BERTOps}}       & \multicolumn{1}{c|}{\textbf{97.37}} & \multicolumn{1}{c|}{\textbf{97.21}} & \multicolumn{1}{c|}{\textbf{97.23}} & \multicolumn{1}{c|}{\textbf{97.97}} & \multicolumn{1}{c|}{\textbf{97.90}} & \multicolumn{1}{c|}{\textbf{97.91}} & \multicolumn{1}{c|}{\textbf{99.38}} & \multicolumn{1}{c|}{\textbf{99.35}} & \textbf{99.36} \\ \hline
\end{tabular}
\label{tab:lfd-results}
\end{table*}
\begin{table*}[tbh]
\centering
% \small
\caption{Comparing baselines with BERTOps on Golden Signal Classification}
\begin{tabular}{|clccccccccc|}
\hline
\multicolumn{11}{|c|}{Golden Signal Classification}                                                                                                                                                                                                                                                                                                                                                                                        \\ \hline
\multicolumn{1}{|c|}{\multirow{2}{*}{Model Type}}           & \multicolumn{1}{c|}{\multirow{2}{*}{Model}} & \multicolumn{3}{c|}{10-shot}                                                                                    & \multicolumn{3}{c|}{20-shot}                                                                                    & \multicolumn{3}{c|}{30-shot}                                                               \\ \cline{3-11} 
\multicolumn{1}{|c|}{}                                      & \multicolumn{1}{c|}{}                       & \multicolumn{1}{c|}{P}              & \multicolumn{1}{c|}{R}              & \multicolumn{1}{c|}{F1}             & \multicolumn{1}{c|}{P}              & \multicolumn{1}{c|}{R}              & \multicolumn{1}{c|}{F1}             & \multicolumn{1}{c|}{P}              & \multicolumn{1}{c|}{R}              & F1             \\ \hline
\multicolumn{1}{|c|}{\multirow{2}{*}{Classical ML}}         & \multicolumn{1}{l|}{Decision Tree}          & \multicolumn{1}{c|}{47.82}          & \multicolumn{1}{c|}{35.46}          & \multicolumn{1}{c|}{36.55}          & \multicolumn{1}{c|}{64.86}          & \multicolumn{1}{c|}{63.36}          & \multicolumn{1}{c|}{63.14}          & \multicolumn{1}{c|}{54.40}          & \multicolumn{1}{c|}{52.72}          & 52.37          \\ \cline{2-11} 
\multicolumn{1}{|c|}{}                                      & \multicolumn{1}{l|}{SGD}                    & \multicolumn{1}{c|}{56.17}          & \multicolumn{1}{c|}{50.35}          & \multicolumn{1}{c|}{50.91}          & \multicolumn{1}{c|}{70.75}          & \multicolumn{1}{c|}{66.90}          & \multicolumn{1}{c|}{67.55}          & \multicolumn{1}{c|}{72.74}          & \multicolumn{1}{c|}{71.16}          & 71.65          \\ \hline
\multicolumn{1}{|c|}{\multirow{5}{*}{Pretrained LLMs}}      & \multicolumn{1}{l|}{ALBERT-base-v2}         & \multicolumn{1}{c|}{54.39}          & \multicolumn{1}{c|}{49.88}          & \multicolumn{1}{c|}{50.02}          & \multicolumn{1}{c|}{66.21}          & \multicolumn{1}{c|}{62.41}          & \multicolumn{1}{c|}{62.97}          & \multicolumn{1}{c|}{78.23}          & \multicolumn{1}{c|}{73.52}          & 74.81          \\ \cline{2-11} 
\multicolumn{1}{|c|}{}                                      & \multicolumn{1}{l|}{ELECTRA}                & \multicolumn{1}{c|}{50.79}          & \multicolumn{1}{c|}{47.75}          & \multicolumn{1}{c|}{48.44}          & \multicolumn{1}{c|}{66.30}          & \multicolumn{1}{c|}{59.34}          & \multicolumn{1}{c|}{59.96}          & \multicolumn{1}{c|}{75.27}          & \multicolumn{1}{c|}{74.70}          & 74.69          \\ \cline{2-11} 
\multicolumn{1}{|c|}{}                                      & \multicolumn{1}{l|}{XLNet}                  & \multicolumn{1}{c|}{55.96}          & \multicolumn{1}{c|}{51.54}          & \multicolumn{1}{c|}{50.38}          & \multicolumn{1}{c|}{69.71}          & \multicolumn{1}{c|}{66.43}          & \multicolumn{1}{c|}{66.60}          & \multicolumn{1}{c|}{76.25}          & \multicolumn{1}{c|}{75.18}          & 75.42          \\ \cline{2-11} 
\multicolumn{1}{|c|}{}                                      & \multicolumn{1}{l|}{RoBERTa}                & \multicolumn{1}{c|}{60.64}          & \multicolumn{1}{c|}{56.03}          & \multicolumn{1}{c|}{55.75}          & \multicolumn{1}{c|}{72.93}          & \multicolumn{1}{c|}{68.32}          & \multicolumn{1}{c|}{69.11}          & \multicolumn{1}{c|}{78.66}          & \multicolumn{1}{c|}{76.60}          & 77.03          \\ \cline{2-11} 
\multicolumn{1}{|c|}{}                                      & \multicolumn{1}{l|}{BERT-base}              & \multicolumn{1}{c|}{61.37}          & \multicolumn{1}{c|}{56.50}          & \multicolumn{1}{c|}{56.03}          & \multicolumn{1}{c|}{\textbf{73.54}} & \multicolumn{1}{c|}{66.19}          & \multicolumn{1}{c|}{67.51}          & \multicolumn{1}{c|}{77.34}          & \multicolumn{1}{c|}{75.89}          & 76.20          \\ \hline
\multicolumn{1}{|c|}{\textbf{Proposed Domain Specific LLM}} & \multicolumn{1}{l|}{\textbf{BERTOps}}       & \multicolumn{1}{c|}{\textbf{62.15}} & \multicolumn{1}{c|}{\textbf{58.39}} & \multicolumn{1}{c|}{\textbf{58.95}} & \multicolumn{1}{c|}{70.90}          & \multicolumn{1}{c|}{\textbf{69.03}} & \multicolumn{1}{c|}{\textbf{69.49}} & \multicolumn{1}{c|}{\textbf{78.51}} & \multicolumn{1}{c|}{\textbf{78.25}} & \textbf{78.30} \\ \hline
\end{tabular}
\label{tab:gs-results}
\end{table*}
\begin{table*}[tbh]
\centering
% \small
\caption{Comparing baselines with BERTOps on Fault Category Prediction}
\begin{tabular}{|clccccccccc|}
\hline
\multicolumn{11}{|c|}{Fault Category Prediction}                                                                                                                                                                                                                                                                                                                                                                                           \\ \hline
\multicolumn{1}{|c|}{\multirow{2}{*}{Model Type}}           & \multicolumn{1}{l|}{\multirow{2}{*}{Model}} & \multicolumn{3}{c|}{10-shot}                                                                                    & \multicolumn{3}{c|}{20-shot}                                                                                    & \multicolumn{3}{c|}{30-shot}                                                               \\ \cline{3-11} 
\multicolumn{1}{|c|}{}                                      & \multicolumn{1}{l|}{}                       & \multicolumn{1}{c|}{P}              & \multicolumn{1}{c|}{R}              & \multicolumn{1}{c|}{F1}             & \multicolumn{1}{c|}{P}              & \multicolumn{1}{c|}{R}              & \multicolumn{1}{c|}{F1}             & \multicolumn{1}{c|}{P}              & \multicolumn{1}{c|}{R}              & F1             \\ \hline
\multicolumn{1}{|c|}{\multirow{2}{*}{Classical ML}}         & \multicolumn{1}{l|}{Decision Tree}          & \multicolumn{1}{c|}{72.38}          & \multicolumn{1}{c|}{46.54}          & \multicolumn{1}{c|}{50.55}          & \multicolumn{1}{c|}{74.41}          & \multicolumn{1}{c|}{47.12}          & \multicolumn{1}{c|}{48.49}          & \multicolumn{1}{c|}{72.87}          & \multicolumn{1}{c|}{52.88}          & 55.38          \\ \cline{2-11} 
\multicolumn{1}{|c|}{}                                      & \multicolumn{1}{l|}{SGD}                    & \multicolumn{1}{c|}{72.04}          & \multicolumn{1}{c|}{49.19}          & \multicolumn{1}{c|}{52.07}          & \multicolumn{1}{c|}{75.85}          & \multicolumn{1}{c|}{59.951}         & \multicolumn{1}{c|}{62.31}          & \multicolumn{1}{c|}{78.31}          & \multicolumn{1}{c|}{62.67}          & 64.94          \\ \hline
\multicolumn{1}{|c|}{\multirow{5}{*}{Pretrained LLMs}}      & \multicolumn{1}{l|}{ALBERT-base-v2}         & \multicolumn{1}{c|}{76.01}          & \multicolumn{1}{c|}{\textbf{62.90}} & \multicolumn{1}{c|}{\textbf{65.85}} & \multicolumn{1}{c|}{77.19}          & \multicolumn{1}{c|}{60.83}          & \multicolumn{1}{c|}{63.73}          & \multicolumn{1}{c|}{78.13}          & \multicolumn{1}{c|}{68.32}          & 69.83          \\ \cline{2-11} 
\multicolumn{1}{|c|}{}                                      & \multicolumn{1}{l|}{ELECTRA}                & \multicolumn{1}{c|}{75.17}          & \multicolumn{1}{c|}{57.95}          & \multicolumn{1}{c|}{62.03}          & \multicolumn{1}{c|}{78.54}          & \multicolumn{1}{c|}{65.32}          & \multicolumn{1}{c|}{68.15}          & \multicolumn{1}{c|}{79.64}          & \multicolumn{1}{c|}{67.28}          & 69.63          \\ \cline{2-11} 
\multicolumn{1}{|c|}{}                                      & \multicolumn{1}{l|}{XLNet}                  & \multicolumn{1}{c|}{75.24}          & \multicolumn{1}{c|}{53.57}          & \multicolumn{1}{c|}{56.82}          & \multicolumn{1}{c|}{76.46}          & \multicolumn{1}{c|}{57.83}          & \multicolumn{1}{c|}{59.83}          & \multicolumn{1}{c|}{80.67}          & \multicolumn{1}{c|}{70.05}          & 72.06          \\ \cline{2-11} 
\multicolumn{1}{|c|}{}                                      & \multicolumn{1}{l|}{RoBERTa}                & \multicolumn{1}{c|}{76.71}          & \multicolumn{1}{c|}{51.73}          & \multicolumn{1}{c|}{53.38}          & \multicolumn{1}{c|}{80.35}          & \multicolumn{1}{c|}{69.70}          & \multicolumn{1}{c|}{71.72}          & \multicolumn{1}{c|}{80.49}          & \multicolumn{1}{c|}{66.82}          & 69.05          \\ \cline{2-11} 
\multicolumn{1}{|c|}{}                                      & \multicolumn{1}{l|}{BERT-base}              & \multicolumn{1}{c|}{76.22}          & \multicolumn{1}{c|}{55.18}          & \multicolumn{1}{c|}{57.94}          & \multicolumn{1}{c|}{78.68}          & \multicolumn{1}{c|}{59.56}          & \multicolumn{1}{c|}{61.69}          & \multicolumn{1}{c|}{80.79}          & \multicolumn{1}{c|}{72.35}          & 74.00          \\ \hline
\multicolumn{1}{|l|}{\textbf{Proposed Domain Specific LLM}} & \multicolumn{1}{l|}{\textbf{BERTOps}}       & \multicolumn{1}{c|}{\textbf{77.33}} & \multicolumn{1}{c|}{50.81}          & \multicolumn{1}{c|}{54.70}          & \multicolumn{1}{c|}{\textbf{79.72}} & \multicolumn{1}{c|}{\textbf{70.16}} & \multicolumn{1}{c|}{\textbf{71.88}} & \multicolumn{1}{c|}{\textbf{81.84}} & \multicolumn{1}{c|}{\textbf{74.65}} & \textbf{76.12} \\ \hline
\end{tabular}
\label{tab:fc-results}
\end{table*}

\subsection{Baseline Models}
In this section, we list the baseline models that we use for comparison in our experiments. LLMs such as BERT are known to outperform LSTMs or RNNs \cite{vaswani2017attention, zhou2023comprehensive}, therefore we do not include them in our study. To answer the first question, we compare the performance of \textit{BERTOps} with state-of-the-art LLMs such as ALBERT-base-v2~\cite{lan2019albert}, ELECTRA~\cite{clark2020electra}, XLNET~\cite{xlnet}, RoBERTa~\cite{roberta}, and BERT-base~\cite{devlin-etal-2019-bert}. 
For most AIOps tasks, the state-of-the-art models are either rule-based or ML-based~\cite{gs_rule, fc_ml}, and to answer the second question, we include two ML models as baselines, Decision Tree(DT) and Stochastic Gradient Descent (SGD). These models were trained using scikit-learn library\footnote{https://scikit-learn.org/stable/} for each of the three downstream tasks.
To answer the third question, we run experiments under the few-shot setting with $k \in \{10, 20, 30\}$ (refer Section~\ref{sec:dataset-prep} and~\ref{sec:finetuning} for details on finetuning).

\subsection{Results}

Tables \ref{tab:lfd-results} - \ref{tab:fc-results} presents weighted Precision, Recall and F1-scores obtained on $10$, $20$ and $30$ shot datasets used for finetuning \textit{BERTOps} on the three downstream tasks. We observe that \textit{BERTOps} outperforms all pretrained LLMs and ML models on the three datasets, except on the Fault Category Prediction when only 10 examples are provided for finetuning. 

For the LFD task, the proposed \textit{BERTOps} LLM learns faster with minimal training data as opposed to other models. Even with as few as $10$ samples per format provided for finetuning, \textit{BERTOps} achieves an F1 score of $97.23\%$. Whereas, when $30$ examples were provided for finetuning, the best ML model (SGD) and the best LLM model(RoBERTa) has a weighted F1-score of $90.34\%$ and $98.3\%$ which is very close to \textit{BERTOps} trained with 10 examples. With $30$ shot dataset, \textit{BERTOps} shows percentage improvement of $9.98\%$ and $1.08\%$ when compared with the best performing ML (SGD) and pre-trained LLM(RoBERTa), respectively.

In the GS Classification task, we observe significant improvements in F1 score with LLMs vis-a-vis the ML models. When the number of training samples per class increase from $20$ to $30$, the best performing ML model (SGD) improves by approximately $4.1$ absolute points (F1 from $67.55$ to $71.65$). Whereas, the best performing LLM model(RoBERTa) and \textit{BERTOps} has an improvement of $7.92$ and $8.81$ absolute points, respectively. These results indicate that \textit{BERTOps} being a domain specific LLM not only outperforms existing state-of-the-art models, but it also gets a good start ($58.95$ vs $50.91$ for SGD and $58.95$ vs $56.03$ for the BERT-base). The percentage improvements of \textit{BERTOps} with respect to the best performing ML(SGD) and LLM (RoBERTa) on $30$ shot dataset are $9.28\%$ and $1.65\%$, respectively.

In the FC Prediction task, \textit{BERTOps} doesn't record a performance gain when finetuned on smaller dataset size of 10 examples, however, it outperforms all baseline methods when finetuned with a larger dataset size of 20 or 30 examples. Also, when finetuning with 10 examples, although ALBERT-base has the highest F1-score, \textit{BERTOps} outperforms all baseline models in terms of precision. This means \textit{BERTOps} exhibited higher precision in identifying the correct fault for a logline, which is an important metric in the FC Prediction task. Precision measures the ratio of number of correct predictions to the total number of predictions. A higher precision means that the BERTOps has less false positives as compared to other methods. On the other hand, Recall measures the ratio of total number of correct predictions to the actual ground truth. A lower recall and a higher precision indicates that BERTOps failed to identify some of the fault category classes, but among the ones that it identified, those were actually correct.  Also, it is impressive to observe that BERTOps quickly learned and corrected itself when provided with more training examples. That is, when 30 training examples were provided for finetuning, \textit{BERTOps} achieves an absolute increase of 23.84 points in recall, which is the highest improvement observed compared to all baseline models. Whereas, ALBERT-base improves only by 5.42 absolute points. We observe a similar trend in F1-score, where BERTOps and ALBERT-base improves by 21.42 and 3.98 absolute points, respectively. 

Also worth noting is the inter-annotator Kappa score and the F1 score of the ML models for the GS and FC classification tasks. The inter-annotator agreement was in the range of 60-65, and the F1 score of ML models for both tasks is in the range of 50-70. Generally, it is believed that inter-annotator agreement places an upper bound on the performance of the ML models\cite{boguslav2017inter}, which is also evident in these results.  Our experiments also indicate an intriguing result where the classical ML models outperforms the state-of-the-art LLM baselines. Specifically, for the GS Classification task with 20-shot dataset for finetuning, the SGD outperforms most of the LLM models such as ALBERT-base, ELECTRA, XLNet, and BERT-base. A similar trend is observed for the FC Prediction task with 20-shot dataset for finetuning, where SGD outperforms both XLNet and BERT. One possible reason for this is the challenges associated with the log data which we also highlighted in the introduction section. Note that, BERT and other LLMs are pre-trained on natural language text, and hence they fail to correctly work with log data with few examples for finetuning on downstream tasks. Because \textit{BERTOps} is pre-trained on log data, it has an obvious advantage in working with log data for the downstream tasks. Moreover, when more training examples are provided to \textit{BERTOps}, it learns quickly and shows significant jumps in performance as opposed to other LLMs. Higher performance of \textit{BERTOps} as compared to both ML and existing LLM models and its ability to learn quickly when more training examples are provided indicate that a LLM for logs is warranted. The experimental findings suggest that \textit{BERTOps} is capable of providing a more accurate generalizable representation for a logline, which can be effectively utilized for various downstream tasks in AIOps.

% Another reason could be that the 10 examples in the training set used for finetuning did not provide sufficient coverage of log variations for each class, which resulted in poor performance of the LLMs compared to ML models. However, when more training examples are provided for finetuning, we observe a signficant improvement in the performance of LLMs. Consequently, the LLMs performed reasonably well in comparison to standard ML models with the increased training data coverage ($30$ shot dataset). The experimental findings suggest that \textit{BERTOps} is capable of providing a high generalizable logline representation, which can be effectively utilized for various downstream tasks in AIOps.

\section{Discussion}

This section presents a detailed analysis of \textit{BERTOps} for each downstream task at the individual class level. The analysis includes a qualitative evaluation of \textit{BERTOps}, finetuned using a 30-shot dataset. In addition, we provide examples of predictions generated by \textit{BERTOps} for the three tasks, juxtaposed with BERT's predictions.

For the LFD task, Table \ref{tab:lfd-confusion} shows a truncated confusion matrix; to save space it is showing only 6 out of 16 classes where predictions are not 100\% accurate, i.e., for the other 10 classes not shown in this table, the predictions were absolutely correct. Among the four held-out log formats (Section \ref{sec:dataset-prep}) - the perfomance on HealthApp, SSH and SendMail is 100\%, while on WebSphere(WS) the performance is 99.18\%. One reason for this superior performance could be the pretraining of \textit{BERTOps} on varying log lines from 17 log formats that enabled it to learn robust and generalized representations. The LFD model is most confused between the following classes: Mac, Spark, and WebSphere. To understand this behavior, Table \ref{tab:qualitative-lfd} analyzes 7 log lines with grounth truth label, along with the predictions of \textit{BERTOps} and BERT-base. 
\begin{table}[!htp]
	\setlength{\tabcolsep}{2pt}
	\centering
% 	\small
    \caption{Confusion matrix for the Log Format Detection task using \textit{BERTOps}; row and column correspond to the true and predicted label, respectively}
    \pgfplotstabletypeset[color cells]
%     {
%     x,Android,Apache,BGL,HDFS,HPC,Hadoop,HA,Mac,OS,Proxifier,SSH,SM,Spark,TB,WS,ZK
%     Android,99.84,0.0,0.0,0.0,0.0,0.0,0.11,0.0,0.0,0.0,0.0,0.0,0.05,0.0,0.0,0.0
%     Apache,0.0,100.0,0.0,0.0,0.0,0.0,0.0,0.0,0.0,0.0,0.0,0.0,0.0,0.0,0.0,0.0
%     BGL,0.0,0.0,100.0,0.0,0.0,0.0,0.0,0.0,0.0,0.0,0.0,0.0,0.0,0.0,0.0,0.0
%     HDFS,0.0,0.0,0.0,100.0,0.0,0.0,0.0,0.0,0.0,0.0,0.0,0.0,0.0,0.0,0.0,0.0
%     HPC,0.0,0.0,0.0,0.0,100.0,0.0,0.0,0.0,0.0,0.0,0.0,0.0,0.0,0.0,0.0,0.0
%     Hadoop,0.0,0.0,0.0,0.0,0.0,99.06,0.0,0.44,0.0,0.0,0.0,0.0,0.15,0.0,0.36,0.0
%     HA,0.0,0.0,0.0,0.0,0.0,0.0,100.0,0.0,0.0,0.0,0.0,0.0,0.0,0.0,0.0,0.0
%     Mac,0.0,0.0,0.0,0.0,0.0,0.0,0.0,98.45,0.0,0.0,0.0,0.0,0.36,0.0,1.19,0.0
%     OS,0.0,0.0,0.0,0.0,0.0,0.0,0.0,0.0,100.0,0.0,0.0,0.0,0.0,0.0,0.0,0.0
%     Proxifier,0.0,0.0,0.0,0.0,0.0,0.0,0.0,0.0,0.0,100.0,0.0,0.0,0.0,0.0,0.0,0.0
%     SSH,0.0,0.0,0.0,0.0,0.0,0.0,0.0,0.0,0.0,0.0,100.0,0.0,0.0,0.0,0.0,0.0
%     SM,0.0,0.0,0.0,0.0,0.0,0.0,0.0,0.0,0.0,0.0,0.0,100.0,0.0,0.0,0.0,0.0
%     Spark,0.0,0.0,0.0,0.0,0.0,1.15,0.0,0.57,0.0,0.0,0.0,0.0,97.13,0.0,1.15,0.0
%     TB,0.0,0.0,0.0,0.0,0.0,0.0,0.0,0.0,0.0,0.0,0.0,0.0,0.0,100.0,0.0,0.0
%     WS,0.34,0.0,0.0,0.0,0.0,0.12,0.34,0.02,0.0,0.0,0.0,0.0,0.0,0.0,99.18,0.0
%     ZK,0.0,0.0,0.0,0.0,0.0,0.0,0.0,0.0,0.0,0.0,0.0,0.0,0.0,0.0,0.0,100.0
% 	}
	{
	 x,          Android,    Hadoop, HealthApp,     Mac,    Spark,  WebSphere
    Android,    99.84,      0.0,    0.11,   0.0,    0.05,   0.0
    Hadoop,     0.0,        99.06,  0.0,    0.44,   0.15,   0.36
    HealthApp,         0.0,        0.0,    100.0,  0.0,    0.0,    0.0
    Mac,        0.0,        0.0,    0.0,    98.45,  0.36,   1.19
    Spark,      0.0,        1.15,   0.0,    0.57,   97.13,  1.15
    WebSphere,         0.34,       0.12,   0.34,   0.02,   0.0,    99.18
	}
	\label{tab:lfd-confusion}
\end{table}

\begin{table}[!htp]
	\setlength{\tabcolsep}{3pt}
	\centering
% 	\small
	\caption{Confusion matrix for the Golden Signal Classification using \textit{BERTOps}, row and column correspond to the true and predicted label, respectively}
    \pgfplotstabletypeset[color cells]{
    x,Availability,Error,Information,Latency,Saturation
    Availability,67.61,21.13,9.86,1.41,0.0
    Error,11.11,77.12,8.5,3.27,0.0
    Information,5.26,6.02,85.71,1.5,1.5
    Latency,6.06,12.12,3.03,78.79,0.0
    Saturation,0.0,18.18,6.06,0.0,75.76
	}
    % \vspace{0.1in}
  	\label{tab:gs-confusion}
\end{table}

\begin{table}[!htp]
	\setlength{\tabcolsep}{2pt}
	\centering
% 	\small
	\caption{Confusion matrix for the Fault Category Prediction task using \textit{BERTOps}, row and column correspond to true and predicted label, respectively. Abbreviated classes are Application(App), Authentication(Auth)}
    \pgfplotstabletypeset[color cells]{
    x,App,Auth,Device,I/O,Memory,Network,Other
    App,76.24,9.9,0.0,0.99,0.0,5.94,6.93
    Auth,0.0,82.61,0.0,0.0,0.0,4.35,13.04
    Device,9.3,2.33,74.42,0.0,6.98,4.65,2.33
    I/O,0.0,0.0,0.0,100.0,0.0,0.0,0.0
    Memory,0.0,0.0,0.0,0.0,100.0,0.0,0.0
    Network,8.54,2.44,0.0,6.1,0.0,73.17,9.76
    Other,10.04,1.94,2.29,0.18,4.93,11.09,69.54
	}
  	\label{tab:fc-confusion}
\end{table}
% Note that, these log lines follow a pattern - an exception type followed by an error message.
% on log lines that are highly similar and follow a pattern of an exception type followed by an error message. %While BERTOps correctly identified 4-7 examples, it misclassified the 1-3. 
While BERT-base misclassified log lines 1-3, \textit{BERTOps} identified subtle nuances in log data to make correct predictions, which is critical for accurately identifying log formats. For example, the term \textit{websphere} appears twice in the example log line number $3$, BERT-base failed to recognize it and misclassified it as of type Spark. Whereas, \textit{BERTOps} predicted the correct label \textit{Websphere} for it. Note that the term \textit{websphere} is not appearing as an independent term in the log line, it is part of a phrase \textit{com.ibm.websphere.security}. Existing pretrained LLMs such as BERT fail to interpret such tokens resulting in wrong predictions.  We mentioned some of the challenges associated with log data in the Introduction Section~\ref{sec:intro}, one such challenge is the lexicon and grammatical structure of log lines. To address these challenges, we preprocessed log data for pretraining the BERTOps model(Section~\ref{sec:dataset-prep}), i.e., using regular expressions to split tokens that are in camelcase format or tokens joined by periods. The preprocessing has helped BERTOps to learn better representation for log lines which is the reason why it identified the log format correctly in these examples. 
These examples demonstrate the ability of \textit{BERTOps} to capture contextual information in log data, which is crucial for precise log format detection.
\begin{table*}[!htp]
\centering
\caption{Qualitative Analysis of BERTOps for Log Format Detection}
\begin{tabular}{|p{12cm}|c|c|c|}
\hline
\multicolumn{1}{|c|}{\textbf{Log Line}} & \textbf{True Labels} & \textbf{BERTOps} & \textbf{BERT-base} \\ \hline
\textbf{1)} Caused by: java.nio.channels.UnresolvedAddressException       & Hadoop & Hadoop & Websphere         \\ \hline

\textbf{2)} WebContainer: 9 ERRORPLUS br.com.plusoft.webapp.PlusoftFilter - javax.servlet.ServletException: org.apache.commons.fileupload.FileUploadBase\$IOFileUploadException: Processing of multipart/form-data request failed. Async operation timed out       & Websphere & Websphere & Hadoop         \\ \hline

\textbf{3)} EST 2/18/20 9:45:46:006 EST com.ibm.websphere.security.PasswordCheckFailedException com.ibm.websphere.security.Cust    & Websphere & Websphere & Spark         \\ \hline

\textbf{4)} RetryExec I org.apache.http.impl.execchain.RetryExec execute I/O exception (java.net.SocketException) caught when processing request to \{s\} -\textgreater \space https://ilesb.southindia.cloudapp.azure.com:443: Connection reset & Websphere & Hadoop & Hadoop \\ \hline

\textbf{5)} java.io.IOException: Failed to create local dir in /opt/hdfs/nodemanager/usercache/curi/appcache /application\_1448006111297\_0137/blockmgr-bcb1cfe2-beb8-4f46-b500-2a6945695a47/04 & Spark & Hadoop & Hadoop \\ \hline 

\textbf{6)} org.apache.avro.AvroTypeException: Attempt to process a enum when a union was expected & Hadoop & Websphere & Hadoop \\ \hline

\textbf{7)} java.io.FileNotFoundException: /opt/hdfs/nodemanager/usercache/curi/appcache/application\_14480\_0137 /blockmgr-518b4be8-5335-426e-8669-d85c295c8087/29/shuffle\_0\_10\_0.index.4ec675d8-4116-4cef-8c74-9a088a849e36 (No such file or directory) & Spark & Hadoop & Hadoop \\ \hline

\end{tabular}
\label{tab:qualitative-lfd}
\end{table*}

\begin{table*}[!htp]
\centering
\caption{Qualitative Analysis of BERTOps for Golden Signal Classification}
\begin{tabular}{|p{12cm}|c|c|c|}
\hline
\multicolumn{1}{|c|}{\textbf{Log Line}} & \textbf{True Labels} & \textbf{BERTOps} & \textbf{BERT-base} \\ \hline
\textbf{1)} 2015-07-29 19:53:10,510 - INFO  [QuorumPeer[myid=3]/0:0:0:0:0:0:0:0:2181:FastLeaderElection@774] - Notification time out: 800	& Latency & Latency & Latency \\ \hline

\textbf{2)} 465980 node-129 unix.hw state\_change.unavailable 1145652795 1 Component State Change: Component is in the unavailable state (HWID=4709) & Availability & Availability & Error \\ \hline

\textbf{3)} 19272 node-100 action error 1100785272 1 boot (command 3314) Error: Unknown We failed to determine the state of this node & Error & Error & Availability \\ \hline

\textbf{4)} 081109 213838 19 WARN dfs.FSDataset: Unexpected error trying to delete block blk\_692167975271429163. BlockInfo not found in volumeMap. & Availability & Error & Error \\ \hline

\textbf{5)} ServletWrappe E com.ibm.ws.webcontainer.servlet.ServletWrapper service Uncaught service() exception thrown by servlet FinGetMessageServlet: java.lang.OutOfMemoryError: PermGen space & Saturation & Error & Error \\ \hline

\textbf{6)} J2CA1010E : Connection not available; timed out waiting for 180 seconds. & Availability & Latency & Latency \\ \hline

\end{tabular}
\label{tab:qualitative-gs}
\end{table*}

\begin{table*}[!htp]
\centering
\caption{Qualitative Analysis of BERTOps for Fault Category Prediction}
\begin{tabular}{|p{11cm}|c|c|c|}
\hline
\multicolumn{1}{|c|}{\textbf{Log Line}} & \textbf{True Labels} & \textbf{BERTOps} & \textbf{BERT-base} \\ \hline

\textbf{1)} WWLM0061W : An error was encountered sending a request to cluster member {CELLNAME=wasp2cell, MEMBERNAME=wasp2b, NODENAME=wasp23anodep2} and that member has been marked unusable for future requests to the cluster & Network & Network & Application \\ \hline

\textbf{2)} Jan  4 13:42:21 LabSZ sshd[32136]: error: Received disconnect from 212.83.176.1: 3: org.vngx.jsch.userauth.AuthCancelException: User authentication canceled by user [preauth] & Authentication & Authentication & Network \\ \hline

\textbf{3)} - 1118787050 2005.06.14 R16-M1-N2-C:J04-U11 2005-06-14-15.10.50.992809 R16-M1-N2-C:J04-U11 RAS KERNEL INFO 1 torus processor sram reception error(s) (dcr 0x02fc) detected and corrected & Device & Memory & Memory \\ \hline

\textbf{4)} 12-17 21:40:04.059   633   692 I TCP     : tcp.conn{4485} TcpClientThread.onRecv(TcpClientThread.java:522) tcp connection event TCP\_CONNECTED & Other & Network & Other \\ \hline

\end{tabular}
\label{tab:qualitative-fc}
\end{table*}

Among the five classes in the GS Classification task, \textit{BERTOps} had the least confusion in identifying the Information class (Table \ref{tab:gs-confusion}). To enhance incident resolution in AIOps, it is crucial to minimize false positives in the informational class as they can impede the fault diagnosis process (Section \ref{sec:downstream-tasks}). This result indicates that \textit{BERTOps} is able to reduce false alarm rate which is very useful for an SRE. 
% and it was most confused in three pairs (Availability, Error), (Latency, Error), and (Saturation, Error), as shown in Table \ref{tab:gs-confusion}. For example, BERTOps missclassified 21.13\% examples as Error whose true label is Availability.

However, \textit{BERTOps} performed relatively poorly in correctly identifying log lines of type Availability, 21.13\% of them are misclassified as Error class. The confusion matrix also shows that \textit{BERTOps} is confusing 12.12\% Latency examples and 18.18\% of Saturation examples with the Error class. To understand the reason for this behavior, we identified some of the examples listed in Table\ref{tab:qualitative-gs}. It shows five log lines, its true labels and predictions from \textit{BERTOps} and BERT. The first three examples demonstrate that \textit{BERTOps} correctly identified golden signals by focusing on domain-specific keywords in the log lines, such as \textit{"time out"} (first example), \textit{"unavailable state"} (second example) and \textit{"action error"} or \textit{"Error: Unknown"} (third example). However, in the fourth example, \textit{BERTOps} might have paid more attention to the phrase \textit{"Unexpected error"} compared to the phrase \textit{"BlockInfo not found"}, leading to an incorrect prediction class Error whereas the actual class is Availability. Similarly, in the fifth example, \textit{BERTOps} might have given more weight to phrase "service() exception", which resulted in misclassification. The predicted class is Error whereas the actual class is Saturation. Ideally, it should have focused on \textit{"java.lang.OutOfMemoryError"}, which is a more indicative signal of the saturation signal. The sixth example is particularly interesting because it contains phrases related to both availability (\textit{Connect not available}) and latency (\textit{timed out waiting}) classes, making it difficult for the model to assign a single golden signal. While the annotators chose availability as the correct label, \textit{BERTOps} predicted it as Latency, which highlights the ambiguity and complexity of this task. This example underscores the difficulty of precisely assigning golden signal labels to log lines that possess subtle nuances and multiple signals, to the point of even confounding human experts.
% Our analysis of Fault Category Prediction task reveals that BERTOps outperforms BERT-base in predicting \textit{application}, \textit{authentication}, \textit{network}, and \textit{other} fault categories, in terms of F1-scores. However, BERTOps exhibits a lower precision score compared to BERT-base in the remaining classes, i.e., \textit{device}, \textit{i/o}, \textit{memory}, indicating more misclassifications of irrelevant instances. Interestingly, our analysis demonstrates that BERTOps excels in identifying fault categories with abnormal behavior\footnote{Abnormal behavior may include errors, warnings, exceptions, and other unexpected behaviors that may indicate a problem or fault in the system} in the log lines but struggles to classify fault categories when there is no such behavior present. Table \ref{tab:qualitative-analysis} provides examples that demonstrate this pattern. To address this issue, one potential solution is to increase the amount of training data to encompass a broader range of variation, leading to more accurate predictions of fault categories.

For the FC Prediction task (Table\ref{tab:fc-confusion}), we found that \textit{BERTOps} was highly accurate in identifying log lines related to memory and I/O faults, and the second-best prediction is observed in authentication (Auth) class. \textit{BERTOps} also demonstrated similar accuracy in identifying application(App), device, and network faults. However, it was confused between the following pairs of class labels, Auth with Other($13.04\%$), Network with App($8.54\%$) and Other($9.76\%$), and Other with App($10.04\%$) and Network($11.09\%$); it appears that the model was confused between the App, Other and Network classes. Table \ref{tab:qualitative-fc} shows examples of some of these cases with true labels along with predictions from \textit{BERTOps} and BERT-base.  
In the first two examples, \textit{BERTOps} correctly identified the fault category by leveraging domain-specific phrases, such as associating phrase \textit{"sending a request"} with class Network in the first example, and associating phrase \textit{"User authentication canceled"} with class Auth in the second example. However, in the third example, BERTOps may have interpreted \textit{SRAM} as a type of memory, leading to a prediction of memory class, whereas the annotators labeled it as device class. This example highlights the complexity of this task, which can be difficult even for a human to classify it. 

The last example related to FC prediction is worth noticing, \textit{BERTOps} excels in identifying fault category in log lines that exhibit abnormal behavior\footnote{Abnormal behavior may include errors, warnings, exceptions, and other unexpected behaviors that may indicate a problem or fault in the system} in log lines, but struggles to classify fault categories when there is no such behavior present. For example, the last example in Table \ref{tab:qualitative-fc} shows that \textit{BERTOps} identified the context (Network), but failed to identify the correct class label Other, this is because the log line is not related to a failure. Based on the above analysis, it can be inferred that while BERTOps performs commendably, and it even outperforms state-of-the-art baselines, there is still scope for improvement in handling the intricacies of AIOps tasks.

\section{Related Work}
\label{sec:related}
Existing LLMs are based on transformer architecture\cite{transformer}, and they can be broadly categorized into three types: encoder-based, decoder-based, and encoder-decoder-based. Encoder-based LLMs consist of an embedding layer followed by a stack of encoder layers, i.e., they use only the encoder layer of the transformer architecture. Examples of encoder-based LLMs are BERT\cite{devlin-etal-2019-bert}, RoBERTa\cite{roberta}, Electra\cite{clark2020electra}, XLNet\cite{xlnet}, etc. Decoder-based LLMs consist of an embedding layer followed by a stack of decoder layers, for example,  GPT3\cite{gpt3}, PaLM\cite{peng-etal-2019-palm}, BLOOM\cite{bloom}, etc. These models are autoregressive at each step, i.e. the input to the model is the output of the previous step, which it uses to generate the output for the following step. Encoder-Decoder-based LLMs consist of both the encoder layer and  the decoder layer.
% The input to the encoder is a sequence of $n$ tokens, and it outputs a sequence of $n$ vectors which is input to a decoder that autoregressively predicts the target sequence. 
Examples of encode-decoder based LLMs are BART\cite{lewis-etal-2020-bart},  T5\cite{t5}, etc.  Encoder-based LLMs are suited for Natural Language Understanding (NLU) tasks, such as classification. The decoder-based and encoder-decoder-based LLMs are suited for generation tasks such as summarization, translation, etc. Our proposed LLM Model for logs is an encoder-based model, this is because the majority of the downstream tasks only need representational information for classifying a log line into a relevant class. 

%Mention about existing papers that they are not truly foundation model. CHange the introduction accordingly.
% LobBERT, BERTLog(https://www.tandfonline.com/doi/full/10.1080/08839514.2022.2145642), UniLog https://arxiv.org/pdf/2112.03159.pdf

Rule-based approaches are commonly applied for the three downstream tasks Log Format Detection, Golden Signal Classification, and Fault Category Detection. Log aggregators like LogDNA\cite{logdna} and Splunk\cite{splunk} use manually curated rules for log format detection and log parsing. Nagar et. al. \cite{9860424 } used a dictionary-based approach that built rules using automatically created dictionaries for each golden signal category. The dictionaries were built using an in-house IT domain-specific Glove word embedding model. Zou et. al. \cite{fault_category} proposed a fault category detection system using manually defined regular expressions and custom built dictionary. However, writing a set of regular expressions can be expensive and difficult to scale and maintain, requiring significant manual effort and engineering. The process of curating new rules for each new log format or previously unseen log can be arduous and time-consuming. 
% Rules may also fail with any log structure or semantic drifts.

 Recently, research works on using transformers for log analysis have emergerd. LogBERT, BERT-Log, NeuralLog are BERT-based models trained specifically for the task of log anomaly detection ~\cite{logbert, bert-log, neurallog}. These models are not truly large language models as they are pre-trained specifically for log anomaly detection task and are not applied to any other downstream tasks. We are the first to propose a LLM for logs, utilizing a transformer architecture and fine-tuning on multiple downstream tasks. 
% \section{Limitations}
% Training a large and deep language model such as BERT requires immense compute resources and takes a lot of time. Additionally, training a LLM requires a huge corpus of log data which is difficult to find since there are not enough resources that are publicly available. One major drawback of domain-specific LLMs is its exclusivity to only one domain. While the model significantly aids in improving accuracy and learning tasks faster in the AIOps domain, it cannot be applied directly to other text domains.
% Due to lack of availability of standard datasets for downstream tasks such as Golden Signal and Fault Category Prediction, in this paper, we had to manually annotate the data. In future, we plan to include more log formats and examples in our corpus with improved Inter Annotator Agreement scores. 

\section{Conclusion and Future Work}
\label{sec:conclusion}
This paper proposes a Transformer-based Large Language Model for AIOps (\textit{BERTOps}) which is trained on a large corpus of public and proprietary log data. We finetuned the pre-trained \textit{BERTOps} model on three downstream log analysis tasks: Log Format Detection, Golden Signal Classification, and Fault Category Prediction. Our experiments show that the proposed \textit{BERTOps} LLM is competitive to both traditional ML and state-of-the-art generic LLM models on the three downstream tasks, that too with minimum training data. And, we observe significant improvements as more training data is provided. For future work, we will apply \textit{BERTOps} on other AIOps tasks such as log parsing, log anomaly detection, incident prediction, and incident prioritization; this will also require defining additional auxiliary tasks for pretraining. In the AIOps domain, there are other modalities of data such as metrics and request traces. We also plan to add data from different modalities during training to produce a holistic and robust \textit{BERTOps} for all AIOps downstream tasks.

\bibliography{custom}

\begin{thebibliography}{10}
\providecommand{\url}[1]{#1}
\csname url@samestyle\endcsname
\providecommand{\newblock}{\relax}
\providecommand{\bibinfo}[2]{#2}
\providecommand{\BIBentrySTDinterwordspacing}{\spaceskip=0pt\relax}
\providecommand{\BIBentryALTinterwordstretchfactor}{4}
\providecommand{\BIBentryALTinterwordspacing}{\spaceskip=\fontdimen2\font plus
\BIBentryALTinterwordstretchfactor\fontdimen3\font minus
  \fontdimen4\font\relax}
\providecommand{\BIBforeignlanguage}[2]{{%
\expandafter\ifx\csname l@#1\endcsname\relax
\typeout{** WARNING: IEEEtran.bst: No hyphenation pattern has been}%
\typeout{** loaded for the language `#1'. Using the pattern for}%
\typeout{** the default language instead.}%
\else
\language=\csname l@#1\endcsname
\fi
#2}}
\providecommand{\BIBdecl}{\relax}
\BIBdecl

\bibitem{loganalysis}
\BIBentryALTinterwordspacing
S.~He, P.~He, Z.~Chen, T.~Yang, Y.~Su, and M.~R. Lyu, ``A survey on automated
  log analysis for reliability engineering,'' \emph{ACM Comput. Surv.},
  vol.~54, no.~6, jul 2021. [Online]. Available:
  \url{https://doi.org/10.1145/3460345}
\BIBentrySTDinterwordspacing

\bibitem{loganalysis1}
D.~Das, M.~Schiewe, E.~Brighton, M.~Fuller, T.~Černý, M.~Bures, K.~Frajták,
  D.~Shin, and P.~Tisnovsky, ``Failure prediction by utilizing log analysis: A
  systematic mapping study,'' 10 2020, pp. 188--195.

\bibitem{mahindru2021log}
R.~Mahindru, H.~Kumar, and S.~Bansal, ``Log anomaly to resolution: Ai based
  proactive incident remediation,'' in \emph{2021 36th IEEE/ACM International
  Conference on Automated Software Engineering (ASE)}.\hskip 1em plus 0.5em
  minus 0.4em\relax IEEE, 2021, pp. 1353--1357.

\bibitem{sre}
\BIBentryALTinterwordspacing
B.~Beyer, C.~Jones, J.~Petoff, and N.~R. Murphy, \emph{Site Reliability
  Engineering: How Google Runs Production Systems}, 2016. [Online]. Available:
  \url{http://landing.google.com/sre/book.html}
\BIBentrySTDinterwordspacing

\bibitem{devlin-etal-2019-bert}
\BIBentryALTinterwordspacing
J.~Devlin, M.-W. Chang, K.~Lee, and K.~Toutanova, ``{BERT}: Pre-training of
  deep bidirectional transformers for language understanding,'' in
  \emph{Proceedings of the 2019 Conference of the North {A}merican Chapter of
  the Association for Computational Linguistics: Human Language Technologies,
  Volume 1 (Long and Short Papers)}.\hskip 1em plus 0.5em minus 0.4em\relax
  Minneapolis, Minnesota: Association for Computational Linguistics, Jun. 2019,
  pp. 4171--4186. [Online]. Available: \url{https://aclanthology.org/N19-1423}
\BIBentrySTDinterwordspacing

\bibitem{gpt3}
\BIBentryALTinterwordspacing
T.~B. Brown, B.~Mann, N.~Ryder, M.~Subbiah, J.~Kaplan, P.~Dhariwal,
  A.~Neelakantan, P.~Shyam, G.~Sastry, A.~Askell, S.~Agarwal, A.~Herbert-Voss,
  G.~Krueger, T.~Henighan, R.~Child, A.~Ramesh, D.~M. Ziegler, J.~Wu,
  C.~Winter, C.~Hesse, M.~Chen, E.~Sigler, M.~Litwin, S.~Gray, B.~Chess,
  J.~Clark, C.~Berner, S.~McCandlish, A.~Radford, I.~Sutskever, and D.~Amodei,
  ``Language models are few-shot learners,'' 2020. [Online]. Available:
  \url{https://arxiv.org/abs/2005.14165}
\BIBentrySTDinterwordspacing

\bibitem{peng-etal-2019-palm}
\BIBentryALTinterwordspacing
H.~Peng, R.~Schwartz, and N.~A. Smith, ``{P}a{LM}: A hybrid parser and language
  model,'' in \emph{Proceedings of the 2019 Conference on Empirical Methods in
  Natural Language Processing and the 9th International Joint Conference on
  Natural Language Processing (EMNLP-IJCNLP)}.\hskip 1em plus 0.5em minus
  0.4em\relax Hong Kong, China: Association for Computational Linguistics, Nov.
  2019, pp. 3644--3651. [Online]. Available:
  \url{https://aclanthology.org/D19-1376}
\BIBentrySTDinterwordspacing

\bibitem{dalle}
\BIBentryALTinterwordspacing
A.~Ramesh, M.~Pavlov, G.~Goh, S.~Gray, C.~Voss, A.~Radford, M.~Chen, and
  I.~Sutskever, ``Zero-shot text-to-image generation,'' 2021. [Online].
  Available: \url{https://arxiv.org/abs/2102.12092}
\BIBentrySTDinterwordspacing

\bibitem{rombach2021highresolution}
R.~Rombach, A.~Blattmann, D.~Lorenz, P.~Esser, and B.~Ommer, ``High-resolution
  image synthesis with latent diffusion models,'' 2021.

\bibitem{10.1109/ICSE-SEIP.2019.00021}
\BIBentryALTinterwordspacing
J.~Zhu, S.~He, J.~Liu, P.~He, Q.~Xie, Z.~Zheng, and M.~R. Lyu, ``Tools and
  benchmarks for automated log parsing,'' in \emph{Proceedings of the 41st
  International Conference on Software Engineering: Software Engineering in
  Practice}, ser. ICSE-SEIP '19.\hskip 1em plus 0.5em minus 0.4em\relax IEEE
  Press, 2019, p. 121–130. [Online]. Available:
  \url{https://doi.org/10.1109/ICSE-SEIP.2019.00021}
\BIBentrySTDinterwordspacing

\bibitem{10.1093/bioinformatics/btz682}
\BIBentryALTinterwordspacing
J.~Lee, W.~Yoon, S.~Kim, D.~Kim, S.~Kim, C.~H. So, and J.~Kang, ``{BioBERT: a
  pre-trained biomedical language representation model for biomedical text
  mining},'' \emph{Bioinformatics}, vol.~36, no.~4, pp. 1234--1240, 09 2019.
  [Online]. Available: \url{https://doi.org/10.1093/bioinformatics/btz682}
\BIBentrySTDinterwordspacing

\bibitem{chalkidis2020legal}
I.~Chalkidis, M.~Fergadiotis, P.~Malakasiotis, N.~Aletras, and
  I.~Androutsopoulos, ``Legal-bert: The muppets straight out of law school,''
  \emph{arXiv preprint arXiv:2010.02559}, 2020.

\bibitem{alsentzer2019publicly}
E.~Alsentzer, J.~R. Murphy, W.~Boag, W.-H. Weng, D.~Jin, T.~Naumann, and
  M.~McDermott, ``Publicly available clinical bert embeddings,'' \emph{arXiv
  preprint arXiv:1904.03323}, 2019.

\bibitem{zhang2020rapid}
R.~Zhang, W.~Yang, L.~Lin, Z.~Tu, Y.~Xie, Z.~Fu, Y.~Xie, L.~Tan, K.~Xiong, and
  J.~Lin, ``Rapid adaptation of bert for information extraction on
  domain-specific business documents,'' \emph{arXiv preprint arXiv:2002.01861},
  2020.

\bibitem{nguyen-etal-2020-bertweet}
\BIBentryALTinterwordspacing
D.~Q. Nguyen, T.~Vu, and A.~Tuan~Nguyen, ``{BERT}weet: A pre-trained language
  model for {E}nglish tweets,'' in \emph{Proceedings of the 2020 Conference on
  Empirical Methods in Natural Language Processing: System
  Demonstrations}.\hskip 1em plus 0.5em minus 0.4em\relax Online: Association
  for Computational Linguistics, Oct. 2020, pp. 9--14. [Online]. Available:
  \url{https://aclanthology.org/2020.emnlp-demos.2}
\BIBentrySTDinterwordspacing

\bibitem{beltagy-etal-2019-scibert}
\BIBentryALTinterwordspacing
I.~Beltagy, K.~Lo, and A.~Cohan, ``{S}ci{BERT}: A pretrained language model for
  scientific text,'' in \emph{Proceedings of the 2019 Conference on Empirical
  Methods in Natural Language Processing and the 9th International Joint
  Conference on Natural Language Processing (EMNLP-IJCNLP)}.\hskip 1em plus
  0.5em minus 0.4em\relax Hong Kong, China: Association for Computational
  Linguistics, Nov. 2019, pp. 3615--3620. [Online]. Available:
  \url{https://aclanthology.org/D19-1371}
\BIBentrySTDinterwordspacing

\bibitem{perplexity}
L.~R. Bahl, F.~Jelinek, and R.~L. Mercer, ``A maximum likelihood approach to
  continuous speech recognition,'' \emph{IEEE Transactions on Pattern Analysis
  and Machine Intelligence}, vol. PAMI-5, no.~2, pp. 179--190, 1983.

\bibitem{he_loghub_2020}
S.~He, J.~Zhu, P.~He, and M.~R. Lyu, ``Loghub: {A} {Large} {Collection} of
  {System} {Log} {Datasets} towards {Automated} {Log} {Analytics},''
  \emph{arXiv:2008.06448 [cs]}, Aug. 2020.

\bibitem{4601543}
Z.~M. Jiang, A.~E. Hassan, P.~Flora, and G.~Hamann, ``Abstracting execution
  logs to execution events for enterprise applications (short paper),'' in
  \emph{2008 The Eighth International Conference on Quality Software}, 2008,
  pp. 181--186.

\bibitem{5936060}
A.~Makanju, A.~N. Zincir-Heywood, and E.~E. Milios, ``A lightweight algorithm
  for message type extraction in system application logs,'' \emph{IEEE
  Transactions on Knowledge and Data Engineering}, vol.~24, no.~11, pp.
  1921--1936, 2012.

\bibitem{8973030}
S.~Messaoudi, A.~Panichella, D.~Bianculli, L.~Briand, and R.~Sasnauskas, ``A
  search-based approach for accurate identification of log message formats,''
  in \emph{2018 IEEE/ACM 26th International Conference on Program Comprehension
  (ICPC)}, 2018, pp. 167--16\,710.

\bibitem{he_drain_2017}
P.~He, J.~Zhu, Z.~Zheng, and M.~R. Lyu, ``Drain: {An} {Online} {Log} {Parsing}
  {Approach} with {Fixed} {Depth} {Tree},'' in \emph{2017 {IEEE}
  {International} {Conference} on {Web} {Services} ({ICWS})}, 2017.

\bibitem{yang2021semi}
L.~Yang, J.~Chen, Z.~Wang, W.~Wang, J.~Jiang, X.~Dong, and W.~Zhang,
  ``Semi-supervised log-based anomaly detection via probabilistic label
  estimation,'' in \emph{2021 IEEE/ACM 43rd International Conference on
  Software Engineering (ICSE)}.\hskip 1em plus 0.5em minus 0.4em\relax IEEE,
  2021, pp. 1448--1460.

\bibitem{fault_category}
D.~Zou, H.~Qin, H.~Jin, W.~Qiang, Z.~Han, and X.~Chen, ``Improving log-based
  fault diagnosis by log classification,'' in \emph{Network and Parallel
  Computing}, C.-H. Hsu, X.~Shi, and V.~Salapura, Eds., 2014.

\bibitem{kappa}
\BIBentryALTinterwordspacing
J.~Cohen, ``A coefficient of agreement for nominal scales,'' \emph{Educational
  and Psychological Measurement}, vol.~20, no.~1, pp. 37--46, 1960. [Online].
  Available: \url{https://doi.org/10.1177/001316446002000104}
\BIBentrySTDinterwordspacing

\bibitem{AdamW}
\BIBentryALTinterwordspacing
I.~Loshchilov and F.~Hutter, ``Decoupled weight decay regularization,'' 2017.
  [Online]. Available: \url{https://arxiv.org/abs/1711.05101}
\BIBentrySTDinterwordspacing

\bibitem{vaswani2017attention}
A.~Vaswani, N.~Shazeer, N.~Parmar, J.~Uszkoreit, L.~Jones, A.~N. Gomez,
  {\L}.~Kaiser, and I.~Polosukhin, ``Attention is all you need,''
  \emph{Advances in neural information processing systems}, vol.~30, 2017.

\bibitem{zhou2023comprehensive}
C.~Zhou, Q.~Li, C.~Li, J.~Yu, Y.~Liu, G.~Wang, K.~Zhang, C.~Ji, Q.~Yan, L.~He
  \emph{et~al.}, ``A comprehensive survey on pretrained foundation models: A
  history from bert to chatgpt,'' \emph{arXiv preprint arXiv:2302.09419}, 2023.

\bibitem{lan2019albert}
Z.~Lan, M.~Chen, S.~Goodman, K.~Gimpel, P.~Sharma, and R.~Soricut, ``Albert: A
  lite bert for self-supervised learning of language representations,''
  \emph{arXiv preprint arXiv:1909.11942}, 2019.

\bibitem{clark2020electra}
K.~Clark, M.-T. Luong, Q.~V. Le, and C.~D. Manning, ``Electra: Pre-training
  text encoders as discriminators rather than generators,'' \emph{arXiv
  preprint arXiv:2003.10555}, 2020.

\bibitem{xlnet}
Z.~Yang, Z.~Dai, Y.~Yang, J.~Carbonell, R.~Salakhutdinov, and Q.~V. Le,
  \emph{XLNet: Generalized Autoregressive Pretraining for Language
  Understanding}.\hskip 1em plus 0.5em minus 0.4em\relax Red Hook, NY, USA:
  Curran Associates Inc., 2019.

\bibitem{roberta}
\BIBentryALTinterwordspacing
L.~Zhuang, L.~Wayne, S.~Ya, and Z.~Jun, ``\BIBforeignlanguage{English}{A
  robustly optimized {BERT} pre-training approach with post-training},'' in
  \emph{\BIBforeignlanguage{English}{Proceedings of the 20th Chinese National
  Conference on Computational Linguistics}}.\hskip 1em plus 0.5em minus
  0.4em\relax Huhhot, China: Chinese Information Processing Society of China,
  Aug. 2021, pp. 1218--1227. [Online]. Available:
  \url{https://aclanthology.org/2021.ccl-1.108}
\BIBentrySTDinterwordspacing

\bibitem{gs_rule}
S.~Nagar, S.~Samanta, P.~Mohapatra, and D.~Kar, ``Building golden signal based
  signatures for log anomaly detection,'' in \emph{2022 IEEE 15th International
  Conference on Cloud Computing (CLOUD)}, 2022, pp. 203--208.

\bibitem{fc_ml}
D.~Zou, H.~Qin, H.~Jin, W.~Qiang, Z.~Han, and X.~Chen, ``Improving log-based
  fault diagnosis by log classification,'' in \emph{Network and Parallel
  Computing: 11th IFIP WG 10.3 International Conference, NPC 2014, Ilan,
  Taiwan, September 18-20, 2014. Proceedings 11}.\hskip 1em plus 0.5em minus
  0.4em\relax Springer, 2014, pp. 446--458.

\bibitem{boguslav2017inter}
M.~Boguslav and K.~B. Cohen, ``Inter-annotator agreement and the upper limit on
  machine performance: Evidence from biomedical natural language processing.''
  \emph{Studies in health technology and informatics}, vol. 245, pp. 298--302,
  2017.

\bibitem{transformer}
A.~Vaswani, N.~Shazeer, N.~Parmar, J.~Uszkoreit, L.~Jones, A.~N. Gomez,
  L.~Kaiser, and I.~Polosukhin, ``Attention is all you need,'' in
  \emph{Proceedings of the 31st International Conference on Neural Information
  Processing Systems}, ser. NIPS'17.\hskip 1em plus 0.5em minus 0.4em\relax Red
  Hook, NY, USA: Curran Associates Inc., 2017, p. 6000–6010.

\bibitem{bloom}
\BIBentryALTinterwordspacing
{BigScience Workshop}, ``{BLOOM} (revision 4ab0472),'' 2022. [Online].
  Available: \url{https://huggingface.co/bigscience/bloom}
\BIBentrySTDinterwordspacing

\bibitem{lewis-etal-2020-bart}
\BIBentryALTinterwordspacing
M.~Lewis, Y.~Liu, N.~Goyal, M.~Ghazvininejad, A.~Mohamed, O.~Levy, V.~Stoyanov,
  and L.~Zettlemoyer, ``{BART}: Denoising sequence-to-sequence pre-training for
  natural language generation, translation, and comprehension,'' in
  \emph{Proceedings of the 58th Annual Meeting of the Association for
  Computational Linguistics}.\hskip 1em plus 0.5em minus 0.4em\relax Online:
  Association for Computational Linguistics, Jul. 2020, pp. 7871--7880.
  [Online]. Available: \url{https://aclanthology.org/2020.acl-main.703}
\BIBentrySTDinterwordspacing

\bibitem{t5}
C.~Raffel, N.~Shazeer, A.~Roberts, K.~Lee, S.~Narang, M.~Matena, Y.~Zhou,
  W.~Li, and P.~J. Liu, ``Exploring the limits of transfer learning with a
  unified text-to-text transformer,'' \emph{J. Mach. Learn. Res.}, vol.~21,
  no.~1, jan 2020.

\bibitem{logdna}
``Automatically parsed logline components,''
  \url{https://docs.mezmo.com/docs/log-parsing}.

\bibitem{splunk}
``Splunk,'' \url{https://docs.splunk.com/}.

\bibitem{9860424}
S.~Nagar, S.~Samanta, P.~Mohapatra, and D.~Kar, ``Building golden signal based
  signatures for log anomaly detection,'' in \emph{2022 IEEE 15th International
  Conference on Cloud Computing (CLOUD)}, 2022, pp. 203--208.

\bibitem{logbert}
H.~Guo, S.~Yuan, and X.~Wu, ``Logbert: Log anomaly detection via bert,'' in
  \emph{2021 International Joint Conference on Neural Networks (IJCNN)}, 2021,
  pp. 1--8.

\bibitem{bert-log}
\BIBentryALTinterwordspacing
S.~Chen and H.~Liao, ``Bert-log: Anomaly detection for system logs based on
  pre-trained language model,'' \emph{Applied Artificial Intelligence},
  vol.~36, no.~1, p. 2145642, 2022. [Online]. Available:
  \url{https://doi.org/10.1080/08839514.2022.2145642}
\BIBentrySTDinterwordspacing

\bibitem{neurallog}
V.-H. Le and H.~Zhang, ``Log-based anomaly detection without log parsing,'' in
  \emph{2021 36th IEEE/ACM International Conference on Automated Software
  Engineering (ASE)}, 2021, pp. 492--504.

\end{thebibliography}
 \bibliographystyle{ieee_trans}

% \bibitem{b1} G. Eason, B. Noble, and I. N. Sneddon, ``On certain integrals of Lipschitz-Hankel type involving products of Bessel functions,'' Phil. Trans. Roy. Soc. London, vol. A247, pp. 529--551, April 1955.
% \bibitem{b2} J. Clerk Maxwell, A Treatise on Electricity and Magnetism, 3rd ed., vol. 2. Oxford: Clarendon, 1892, pp.68--73.
% \bibitem{b3} I. S. Jacobs and C. P. Bean, ``Fine particles, thin films and exchange anisotropy,'' in Magnetism, vol. III, G. T. Rado and H. Suhl, Eds. New York: Academic, 1963, pp. 271--350.
% \bibitem{b4} K. Elissa, ``Title of paper if known,'' unpublished.
% \bibitem{b5} R. Nicole, ``Title of paper with only first word capitalized,'' J. Name Stand. Abbrev., in press.
% \bibitem{b6} Y. Yorozu, M. Hirano, K. Oka, and Y. Tagawa, ``Electron spectroscopy studies on magneto-optical media and plastic substrate interface,'' IEEE Transl. J. Magn. Japan, vol. 2, pp. 740--741, August 1987 [Digests 9th Annual Conf. Magnetics Japan, p. 301, 1982].
% \bibitem{b7} M. Young, The Technical Writer's Handbook. Mill Valley, CA: University Science, 1989.
% \end{thebibliography}
% \vspace{12pt}
% \color{red}
% IEEE conference templates contain guidance text for composing and formatting conference papers. Please ensure that all template text is removed from your conference paper prior to submission to the conference. Failure to remove the template text from your paper may result in your paper not being published.

\end{document}